\documentclass[runningheads]{llncs}
\usepackage{graphicx}

\usepackage{amsmath,amssymb,amsfonts,graphicx} 
\usepackage{color}
\usepackage{multirow}
\usepackage{setspace}
\usepackage{tocloft}
\usepackage{hyperref}
\usepackage{caption}
\usepackage{subcaption}
\usepackage{wrapfig}
\usepackage{enumitem}

\usepackage{bm}
\begin{document}
\title{Can representation learning for multimodal image registration be improved by supervision of intermediate layers?}
\titlerunning{CoMIRs with Supervised Intermediate Layers}
%
\author{Elisabeth Wetzer\orcidID{0000-0002-0544-8272} \and
Joakim Lindblad\orcidID{0000-0001-7312-8222} \and
Nata\v{s}a Sladoje\orcidID{0000-0002-6041-6310}}
\authorrunning{E. Wetzer et al.}
%
\institute{Centre for Image Analysis, Dept. of Information Technology, Uppsala University,  Uppsala, Sweden
\email{elisabeth.wetzer@it.uu.se}\\
}
\maketitle              
\begin{abstract}
Multimodal imaging and correlative analysis typically require image alignment. Contrastive learning can generate representations of multimodal images, reducing the challenging task of multimodal image registration to a monomodal one. Previously, additional supervision on intermediate layers in contrastive learning has improved biomedical image classification. We evaluate if a similar approach improves representations learned for registration to boost registration performance.
We explore three approaches to add contrastive supervision to the latent features of the bottleneck layer in the U-Nets encoding the multimodal images and evaluate three different critic functions. Our results show that representations learned without additional supervision on latent features perform best in the downstream task of registration on two public biomedical datasets. We investigate the performance drop by exploiting recent insights in contrastive learning in classification and self-supervised learning. We visualize the spatial relations of the learned representations by means of multidimensional scaling, and show that additional supervision on the bottleneck layer can lead to partial dimensional collapse of the intermediate embedding space.

\end{abstract}
%
%
\section{Introduction}
Multimodal imaging enables capturing complementary information about a sample, essential for a large number of diagnoses in digital pathology. However, directly co-aligned data can only be provided if an imaging device hosts multiple imaging modalities, otherwise individually acquired data have to be registered by image processing. Different sensors may produce images of very different appearance, making automated multimodal registration a very challenging task. Consequently, the registration is often performed manually; a difficult, labor- and time consuming, and hence expensive approach. Reliable automated multimodal registration can reduce the workload, allowing for larger datasets to be studied in research and clinical settings.
\begin{figure}[t]
\centering
\includegraphics[width=\textwidth]{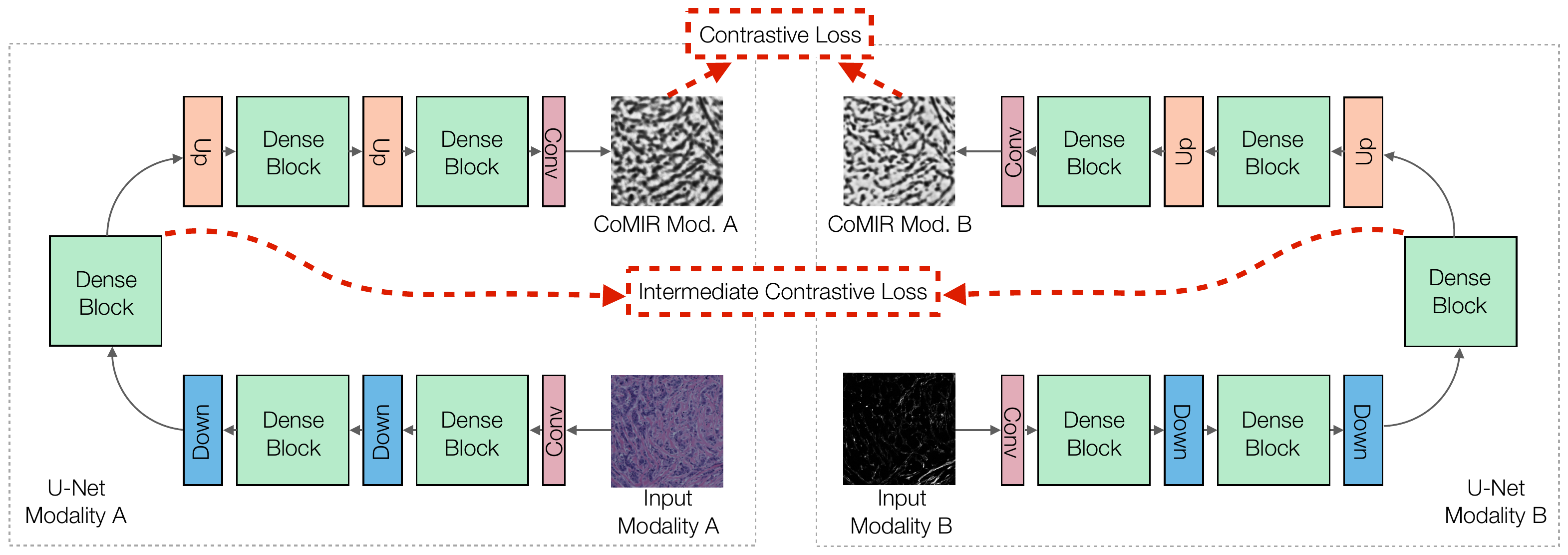}
\caption{Overview of the representation learning approach: Two U-Nets sharing no weights are trained in parallel to learn dense representations (CoMIRs) through a contrastive loss on the final output layers of the networks. Additional contrastive losses acting on the features of the bottleneck (BN) layers are evaluated with an aim to improve CoMIRs for registration.}
\label{fig:CLfig}
\end{figure}
Numerous methods are available for automated monomodal registration, however, multimodal image registration is both more difficult and with fewer tools available. Recently, a modality-independent representation learning technique was introduced in \cite{pielawski2020comir}; the method generates Contrastive Multimodal Image Representations (CoMIRs) for Registration, which can be registered using monomodal (intensity-based and feature-based) registration methods. 
CoMIR-based multimodal registration~\cite{pielawski2020comir} has been successfully applied to various datasets: brightfield (BF) \&  second harmonic generation (SHG) images in~\cite{pielawski2020comir}, remote sensing images (RGB \& near-infrared), and correlative time-lapse quantitative
phase images (QPI) \& fluorescence microscopy (FM) images, and magnetic resonance images (MRI T1 \& T2) in \cite{10.1371/journal.pone.0276196,OFVERSTEDT2022196}. 


Based on a recent study~\cite{NEURIPS2021_c9f06258} which has shown that representations learned contrastively performed better in the downstream task of biomedical image classification when additional contrastive losses were used to supervise intermediate layers in the network, we investigate if such additional supervision can be applied to further improve CoMIRs for the downstream task of registration. 
A schematic overview of the considered approach is given in Fig.~\ref{fig:CLfig}.

Our contributions are as follows: (i) We evaluate three approaches of including contrastive losses on U-Net bottleneck (BN) features when generating CoMIRs of BF and SHG images; for each we test three different similarity/distance measures. 
We observe that \textit{leaving intermediate features unconstrained} results in CoMIRs leading to more successful rigid multimodal image registration. This differs from what was previously observed for features used in biomedical image classification. (ii) We confirm that unconstrained intermediate features result in better registration on a second dataset of QPI and FM images, using the best performing similarity measure observed on the BF and SHG dataset in each approach. (iii) We investigate the reasons for the drop in registration performance and observe that contrastive training on the BN features can lead to a partial dimensional collapse of the feature space. (iv) Furthermore, we evaluate the relation between the quality of the generated representations and the downstream task of image registration, using several image distance/similarity measures.

\section{Background}

\subsection{Representation Learning}
Several recent approaches find common representations for images of different modalities to enable monomodal registration methods. 
The common representation may be one of the involved modalities, e.g., learned by GAN-based Image-To-Image (I2I) translation, where a generator and discriminator are trained in a zero-sum game, generating representations in line with the appearance and statistical distribution of one modality given an image of another modality~\cite{https://doi.org/10.48550/arxiv.2204.13656,10.1007/978-3-030-20351-1_19}.
Another strategy is to map both of the considered image modalities into a common space: 
In \cite{pielawski2020comir}, the 
contrastive loss InfoNCE~\cite{DBLP:journals/corr/abs-1807-03748} was used to produce dense, image-like representations -- called CoMIRs -- which proved useful for multimodal image registration on a variety of different datasets \cite{pielawski2020comir,10.1371/journal.pone.0276196,OFVERSTEDT2022196}. CoMIRs are generated by two U-Nets trained in parallel, sharing no weights, connected by the contrastive loss InfoNCE \cite{hjelm2018learning,pmlr-v9-gutmann10a}, which maximizes a lower bound of mutual information (MI).
Contrastive Learning is also used in \cite{10.1007/978-3-030-58545-7_19} to perform I2I, which inspired the development of ContraReg \cite{https://doi.org/10.48550/arxiv.2206.13434} to contrastively learn to perform unsupervised, deformable multimodal image registration.

A number of I2I approaches \cite{isola2017image,CycleGAN2017,9157662,drti} were evaluated in \cite{pielawski2020comir,10.1371/journal.pone.0276196} 
for multimodal registration, but the representations lacked similarity in structures or intensity needed for registration.
CoMIRs \cite{pielawski2020comir}, however, performed well in combination with both intensity-based \cite{8643403} and with feature-based registration methods.

\smallskip
\noindent
\textbf{Additional Supervision on Intermediate Layers}\\
Recently it was shown in \cite{NEURIPS2021_c9f06258} that a momentum contrastive (MoCo) learning-based framework benefits from additional supervision imposed on intermediate layers. The features are more similar earlier in the network, resulting in better performance in biomedical image classification.
In \cite{8451804}, a three stream architecture (TS-Net) is proposed combining Siammese and pseudo-Siammese networks for patch matching of multimodal images. An additional contrastive loss on intermediate features improves the matching performance on three multimodal datasets.
Additional supervision on intermediate layers in U-Nets has mainly evaluated on the BN layer. In \cite{LI2020113131}, additional supervision on the BN features is provided for liver and tumor segmentation. The authors argue that supervision on the BN layer can reduce information loss, due to its highly condensed low-dimensional information of label maps. In \cite{9481244}, a Tunable U-Net (TU-Net) uses a tuning parameter acting on the BN features, to change the output without the requirement of multiple trained models. In \cite{9162061}, 
the effect of fine-tuning different sets of layers of a pretrained U-Net for ultrasound and X-ray image segmentation is studied. The authors show that the choice of layers is critical for the downstream tasks and differs significantly between segmentation and classification. 
This is based on the assumption that low-level features, which are associated with shallow layers, are shared across different data sets. The authors observe that freezing the BN has equivalent segmentation performance as fine-tuning the entire network, highlighting the importance of that particular latent representation.

Similarly, we focus on evaluating additional losses constraining the BN layer in the U-Nets which generate CoMIRs for alignment of multimodal images.
\subsection{Contrastive Learning}
\vspace{-5pt}
Contrastive Learning (CL) has been successfully employed in many tasks - from self-supervised pretraining, to image retrieval and classification \cite{DBLP:journals/corr/abs-1906-05849,8451804,pielawski2020comir,10.1371/journal.pone.0276196,NEURIPS2020_4c2e5eaa,https://doi.org/10.48550/arxiv.2206.13434,NEURIPS2021_c9f06258}. The goal of CL is to learn a representation s.t. similar samples are mapped close to each other, while dissimilar samples are not. \textit{Similar} can refer to the same class, or different views or augmentations of one particular sample. Usually, the learned representation is a 1D vector. 
CoMIRs differ in that they are 2D image-like representations intended to preserve spatial structures, not only optimized for their distinction to other samples. Learning CoMIRs is closely related to self-supervised learning (SSL). While CoMIR training currently relies on a labelled training set in form of aligned image pairs, the two images in different modalities of one sample act as different views of one sample in SSL.

The features of multiple views in SSL depend strongly on data augmentation (DA), as it decouples the correlations of features between the representations of positive samples~\cite{pmlr-v139-wen21c}. The importance of view selection is also addressed in \cite{NEURIPS2020_4c2e5eaa,xiao2021what}. The authors in \cite{NEURIPS2020_4c2e5eaa} find that the MI between views should be reduced while information relevant for the downstream task has to be preserved. 
While \cite{pmlr-v139-wen21c,NEURIPS2020_4c2e5eaa} argue for extensive DA, in \cite{https://doi.org/10.48550/arxiv.2205.06926} it was shown that extensive DA can yield the projector head of a SSL network invariant to the DA to a higher degree than the encoder itself, resulting in a projection to a low-dimensional embedding space. 
This so-called \textit{dimensional collapse} (DC) has been observed in different CL settings \cite{Jing2021UnderstandingDC,https://doi.org/10.48550/arxiv.2205.06926,https://doi.org/10.48550/arxiv.2204.09179,Hua_2021_ICCV,Wang_2021_CVPR} and is currently subject of research. The phenomenon has been first studied in \cite{Jing2021UnderstandingDC} and \cite{Hua_2021_ICCV}. 
DC can occur if the variance caused by DA is larger than the variance in the data distribution, implicit regularization to favor low-rank solutions, or strong overparametrization~\cite{Jing2021UnderstandingDC}. For non-contrastive SSL methods such as SimSiam or BYOL, underparametrization of the model can cause at least partial collapse in \cite{10.1007/978-3-031-19821-2_28}.

Following these recent observations, we inspect if DC of the feature space is the cause of representations which are unsuitable for feature-based registration.

\section{Method}
\label{sec:method}
CoMIRs, as originally proposed, are learned by two U-Nets~\cite{jegou2017one}, sharing no weights, connected by a contrastive loss given as follows:

For $\mathcal{D} = \{(\bm{x}_i^1, \bm{x}_i^2)\}_{i=1}^n$ an i.i.d. dataset containing $n$ data points, $\bm{x}^j$ is an image in modality $j$, and $f_{\bm{\theta}_j}$ the network processing modality $j$ with respective parameters $\bm{\theta}_j$ for $j \in \{1,2\}$.
Given an arbitrary image pair $\bm{x}=(\bm{x}^1, \bm{x}^2) \in \mathcal{D}$, the loss is given by
\vspace{-5pt}
\begin{align}
    & \mathcal{L}(\mathcal{D}) = 
    - \frac{1}{n}\sum_{i=1}^{n}\left(
        \log \frac{
            e^{h(\bm{y}^1_i,\bm{y}^2_i)/\tau}
        }{
            e^{h(\bm{y}^1_i,\bm{y}^2_i)/\tau} +
            \sum_{j \neq i} e^{h(\bm{y}^1_j,\bm{y}^2_j)/\tau}
        }\right)\,.
    \label{eq:infonce_critic}
\end{align}
$\mathcal{L}(\mathcal{D})$ is named InfoNCE as described in \cite{oord2018representation}.
The exponential of a similarity function (called critic) $h(\bm{y}^1,\bm{y}^2)$ computes a chosen similarity between the representations $\bm{y}^1=f_{\bm{\theta}_1}(\bm{x}^1)$ and $\bm{y}^2=f_{\bm{\theta}_2}(\bm{x}^2)$ for the scaling parameter $\tau>0$. 

\subsection{Additional Supervision of the BN latent representation}
\label{sec:addedloss}
\subsubsection{Different Critic functions}
We consider three types of similarity functions in Eqn. \ref{eq:infonce_critic} for the supervision of the BN latent representations:
\begin{itemize}[noitemsep,topsep=0pt]
 \setlength\itemsep{0.1em}
    \item A Gaussian model with a constant variance $h(\bm{y}^1,\bm{y}^2) = -||\bm{y}^1 - \bm{y}^2||_2^2$ which uses the $L^2$ norm, i.e. mean squared error (MSE) as a similarity function; 
    \item A trigonometric model $h( \bm{y}^1,\bm{y}^2) = \frac{\langle \bm{y}^1,\bm{y}^2 \rangle}{||\bm{y}^1||~||\bm{y}^2||}$ relating to cosine similarity; 
    \item A model using the $L^1$ norm as a similarity $h(\bm{y}^1,\bm{y}^2) = -||\bm{y}^1 - \bm{y}^2||_1$.
\end{itemize}
\vspace{10pt}
\noindent
We investigate the following approaches for supervision of the BN:

\textbf{Approach I: Alternating loss}
$\mathcal{L}(\mathcal{D})$ given in Eq. \eqref{eq:infonce_critic} is computed in an alternating way on the final output representations in the network as $\mathcal{L}_{C}(\mathcal{D})$ and on the BN latent features as $\mathcal{L}_{BN}(\mathcal{D})$, taking turns every iteration. A hyperparameter ensures losses of the same magnitude for stable training.

\textbf{Approach II: Summed loss}
As proposed in \cite{NEURIPS2021_c9f06258}, the loss in Eq.~\eqref{eq:infonce_critic} is calculated in each iteration on the final representations in the network and on the BN latent features. The two losses are combined in a weighted sum, requiring an additional hyperparameter $\alpha$ to ensure that the two losses are of the same magnitude: $\mathcal{L}_{Sum}(\mathcal{D})=\mathcal{L}_{C}(\mathcal{D})+\alpha \mathcal{L}_{BN}(\mathcal{D})$

\textbf{Approach III: Pretraining using a contrastive loss on the BN layers}
For this approach, the networks are trained with $\mathcal{L}_{BN}(\mathcal{D})$ for the first 50 epochs on the BN latent features. After this pretraining of the networks, $\mathcal{L}_{C}(\mathcal{D})$ is computed acting only on the final layer for 50 more epochs.

\subsection{Implementation Details}
All models are trained for 100 epochs, except the baseline model denoted "Baseline 50" in Fig.~\ref{fig:barplots}, which is trained for 50 epochs. In all experiments, $\mathcal{L}_{C}(\mathcal{D})$ uses MSE as $h(\bm{y}^1,\bm{y}^2)$ (as suggested in \cite{pielawski2020comir}), while the similarity functions acting on the BN layers are varied between none, MSE, cosine similarity, and $L^1$-norm as described in Sect.~\ref{sec:addedloss}. In all cases 1-channel CoMIRs are generated with identical, random data augmentation consisting of random flips, rotations, Gaussian blur, added Gaussian noise and contrast adjustments. More detailed implementation information, including all chosen hyperparameters, can be found in appendix Sec. \ref{appendix:implementation}.

\section{Evaluation}

\subsection{Datasets}

\textbf{SHG \& BF Dataset}\\
The dataset is a publicly available, manually aligned registration benchmark \cite{kevin_eliceiri_2020_3874362} consisting of SHG and BF crops of tissue micro-array (TMA) image pairs. 
The training set consists of 40 image pairs of size $834 \times 834$\,px cropped from original TMA images \cite{fullcores} to avoid any border affects due to  transformations. 
The test set comprises 134 pairs of synthetically, randomly rotated and translated images of up to ±30 degrees, and ±100\,px.

\smallskip
\noindent
\textbf{QPI \& FM Dataset}\\
The dataset consists of simultaneously acquired correlative time-lapse QPI \cite{tomas_vicar_2019_2601562} and FM images \cite{vicar_tomas_2021_4531900} of three prostatic cell lines, captured as time-lapse stacks at seven different fields of view while exposed to cell death inducing compounds. It is openly available for multimodal registration~\cite{lu_jiahao_2021_5557568} and is used as in \cite{10.1371/journal.pone.0276196}. The images are of size $300 \times  300$\,px, with 420 test samples in each of three evaluation folds. All images originating from one cell line are used as one fold in 3-folded cross-validation. The test set was created by synthetic random rotations of up to ±20 degrees and translations of up to ±28\,px. 

\subsection{Evaluation Metrics}
\textbf{Registration performance:} 
We register CoMIRs by extracting Scale-Invariant Feature Transform (SIFT, \cite{sift}) features, and match them by Random Sample Consensus (RANSAC \cite{ransac}). 
The registration error is calculated as $err =\frac{1}{4} \sum_{i=1}^4 ||C_i^{Ref}-C_i^{Reg}||_2$,
where $C_i$ are the corner points of the reference image $C_i^{Ref}$, and the registered image $C_i^{Reg}$ respectively.
The registration success rate (RSR) is measured by the percentage of test images which are successfully registered, whereas success is defined by an error below a certain threshold. 
The implementation details, including parameter choices, are reported in appendix Sec. \ref{appendix:implementation}.

\smallskip
\noindent
\textbf{Measuring Representation Quality:}
\label{sec:img_metrics}
Intuitively, the higher similarity in appearance the images have, the more successful their registration. 
Therefore, the ``goodness'' of CoMIRs can be evaluated in two ways: by evaluating (i) the similarity/distance between the CoMIRs of the corresponding images; and (ii) the success of their registration. We correlate both types of evaluation in this study. For comparing CoMIRs we utilize several approaches common for quantifying image similarity/distance. More precisely, we evaluate the following:
\begin{itemize}[noitemsep,topsep=0pt]
    \item the pixelwise measures \textbf{MSE} and \textbf{correlation},
    \item the perceptual similarity measure \textbf{structural similarity index measure} (SSIM, \cite{1284395}), 
    \item a distance measure which combines intensity and spatial information, namely \textbf{$\mathbf{\alpha}$-Average Minimal Distance} ($\alpha$-AMD, \cite{8643403,6645422}), 
    \item a distance measure comparing two distributions and is a popular choice to evaluate GAN generated representations~\cite{morozov2021on}, namely \textbf{Fréchet Inception Distance} (FID, \cite{NIPS2017_8a1d6947}).
\end{itemize}
The respective definitions are given in appendix Sec. \ref{appendix:measures}.

\section{Results}
\label{sec:results}
\begin{figure}[h]
\def\maxH{3.6cm}
\centering
     \begin{subfigure}[t]{0.37\linewidth}%
         \centering
         \vtop to \maxH{\null\includegraphics[trim=25 0 52 25, clip = true, width=\textwidth, height=\maxH, keepaspectratio]{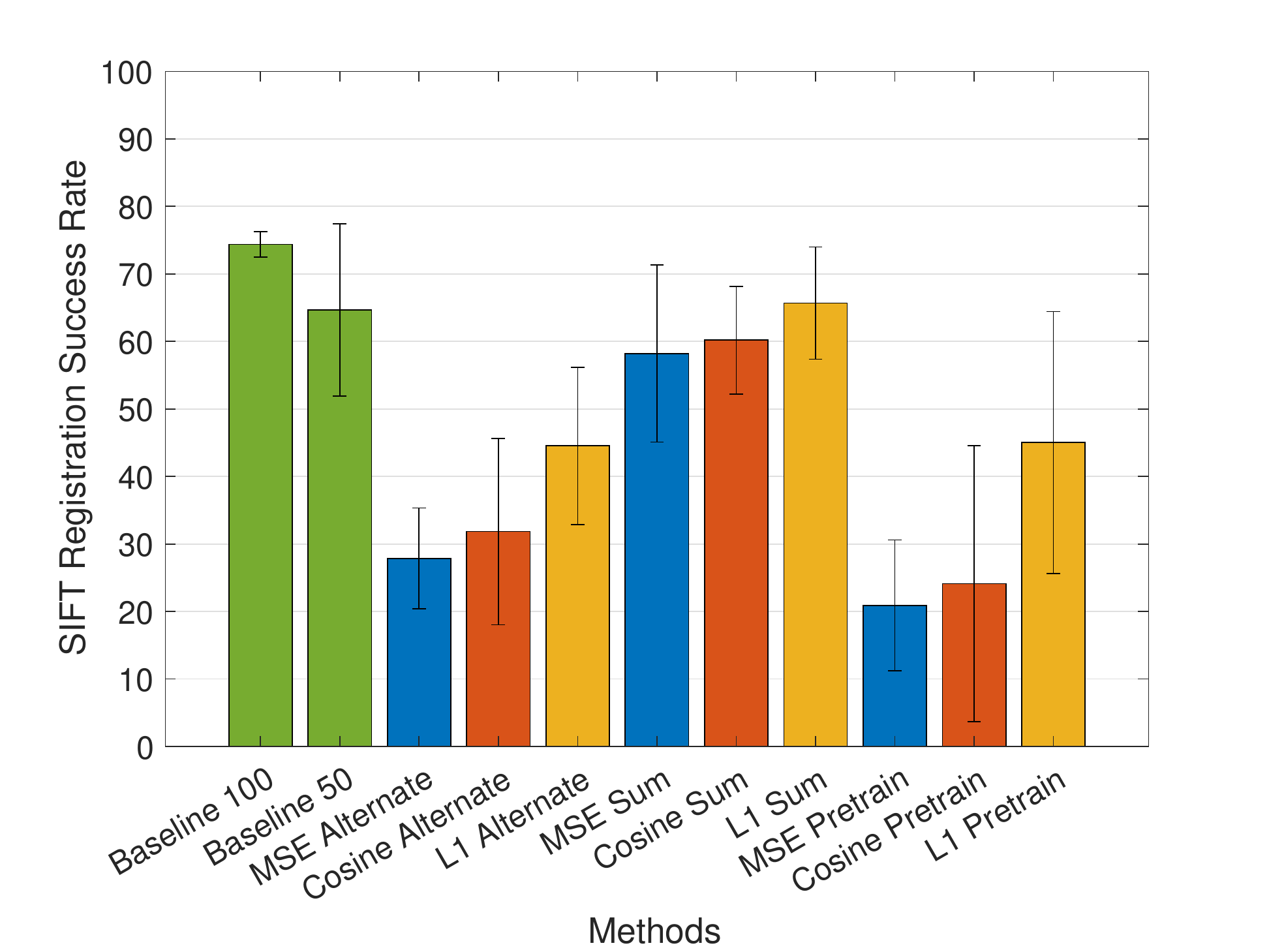}\vfill}
         \subcaption{RSR for BF \& SHG}
         \label{fig:SIFTBarplot}
     \end{subfigure}%
\quad
\quad
     \begin{subfigure}[t]{0.32\textwidth}%
        \centering
         \vtop to \maxH{\null\includegraphics[trim=25 0 52 25, clip = true, width=\textwidth, height=\maxH, keepaspectratio]{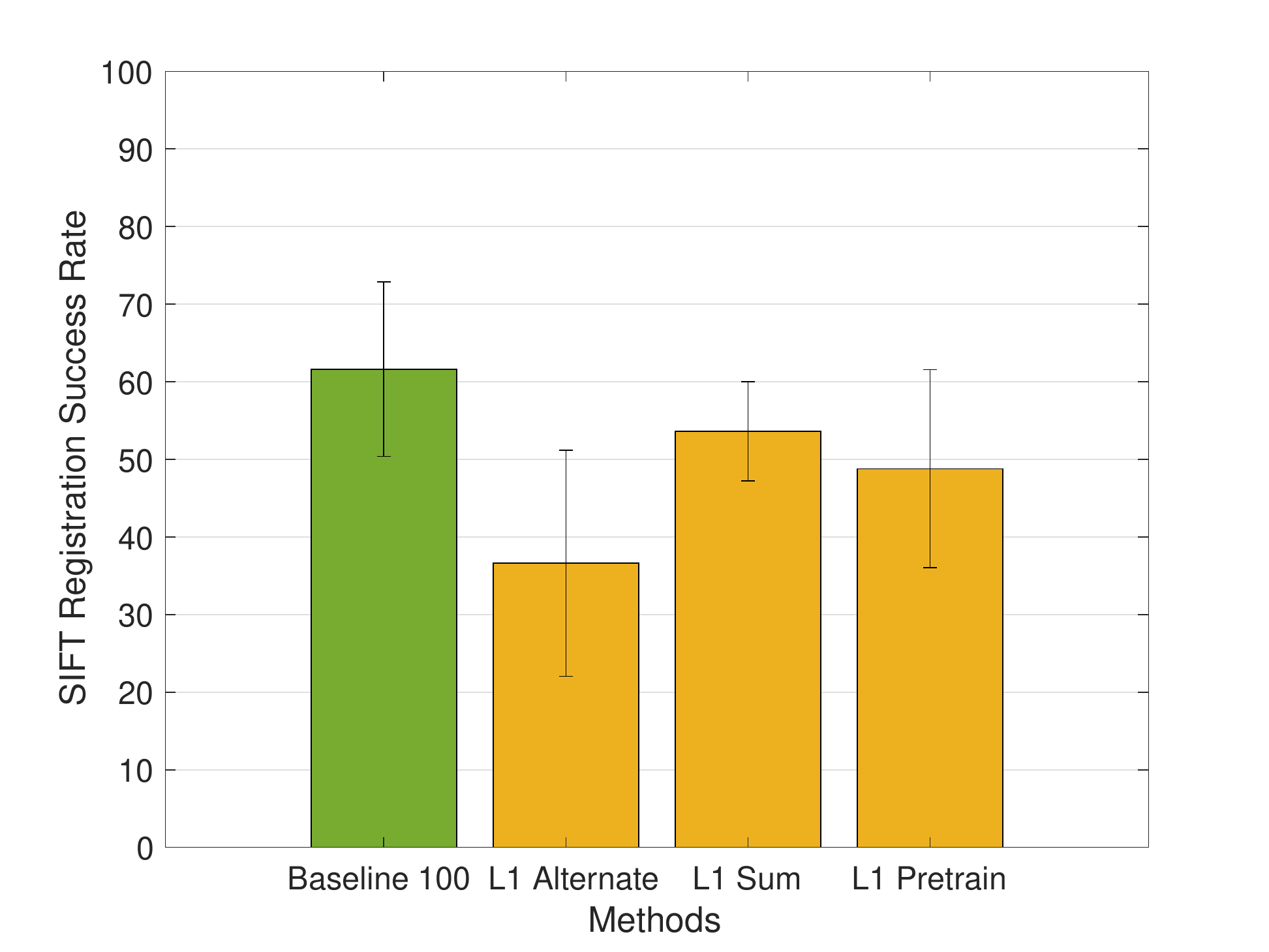}\vfill}
         \subcaption{RSR for QPI \& FM}
         \label{fig:CytoBarplot}     
         \end{subfigure}%
            \caption{Evaluation of the CoMIRs w.r.t.\ to registration performance.}
        \label{fig:Cyto}
\end{figure}

\textbf{Registration Performance}\\
\noindent
Fig.~\ref{fig:SIFTBarplot} shows the RSR, computed as the percentage of the test set which was registered with an error less than 100\,px on the BF \& SHG dataset. The methods are grouped color-wise w.r.t. to the similarity function used for the BN supervision. In green the baseline results are shown for CoMIRs as introduced in \cite{pielawski2020comir}, using no additional loss on intermediate layers. Detailed results of each run per experiment and examples of CoMIRs for the tested approaches can be found in appendix Sec. \ref{appendix:results}.

It should be considered that the Approaches I and III only spend half of all epochs on evaluating $\mathcal{L}_{C}(\mathcal{D})$, ensuring pixel-wise intensity similarity in the final CoMIRs. To confirm that this does not cause the subsequent lower registration performance, a second set of baseline experiments was run for only 50 instead of 100 epochs. 
Hence, the plots include the baseline trained for 100 epochs ("Baseline 100") and also for 50 epochs ("Baseline 50").

The best performing critic function, w.r.t. the registration performance on the BF \& SHG Dataset (i.e. $L^1$), is used to assess the three approaches 
on the second multimodal dataset, containing QPI and FM images of cells. This dataset confirms our observations that no supervision on intermediate layers results in CoMIRs more suitable for registration. 
The RSR (averaged over three folds) corresponding to the percentage of the test set with a relative registration error of less than 2\% of the image size, as used in \cite{10.1371/journal.pone.0276196}, is shown in Fig.~\ref{fig:CytoBarplot} (detailed results are listed in the appendix). 
We observe that on both datasets the Baseline approach outperforms the considered alternatives, producing CoMIRs which lead to the highest RSR. 

\smallskip
\noindent
\textbf{Representation Quality}\\
We report the image similarities/distances as introduced in Sec. \ref{sec:img_metrics} for CoMIRs produced on the BF \& SHG dataset. 
Table~\ref{table} lists the performance measures computed for the test set (median over all images where applicable), averaged over three runs.
\begin{table}[t]
\caption{The medians of all the considered performance measures, computed on the test set and  averaged over three runs, for the different approaches to include supervision of the BN latent representations. Arrows indicate if a high (\textuparrow) or low (\textdownarrow) value correspond to a good result.}
\centering
\label{table}
\smallskip\resizebox{0.95\textwidth}{!}{
\begin{tabular}{ll|rrrrrr}
Approach  & Intermediate Loss & MSE\,\textdownarrow      & SSIM\,\textuparrow & Correlation\,\textuparrow & $\alpha$-AMD\,\textdownarrow & \;\;FID\,\textdownarrow    & Reg. Success Rate\,\textuparrow \\ \hline
baseline 100  & none              & 4,771 & 0.53 & \textbf{0.66}        & 1.86     & \textbf{93.26}  & \textbf{74.38}    \\
alternate & MSE               & 4,534 & 0.46 & 0.36        & 2.21     & 137.89 & 27.86             \\
alternate & Cosine            & 3,949 & 0.50 & 0.42        & \textbf{1.32}     & 123.31 & 31.84             \\
alternate & L1                & 7,603 & 0.37 & 0.47        & 5.68     & 190.26 & 44.53             \\
sum       & MSE               & 4,674 & \textbf{0.57} & 0.60        & 2.33     & 129.31 & 58.21             \\
sum       & Cosine            & 5,776 & 0.48 & 0.52        & 7.35     & 219.76 & 60.20             \\
sum       & L1                & \textbf{3,894} & \textbf{0.57} & 0.64        & 1.70     & 106.09 & 65.67             \\
pretrain  & MSE               & 4,815 & 0.37 & 0.32        & 3.46     & 179.98 & 20.90             \\
pretrain  & Cosine            & 8,139 & 0.35 & 0.28        & 7.31     & 181.93 & 24.13             \\
pretrain  & L1                & 5,668 & 0.44 & 0.48        & 2.86     & 160.34 & 45.02            
\end{tabular}
}
\vspace{-10pt}
\end{table}
\begin{figure}[h]
     \centering
     \begin{subfigure}[t]{0.32\textwidth}
         \centering
         \includegraphics[width=\textwidth,trim=13 8 5 0, clip]{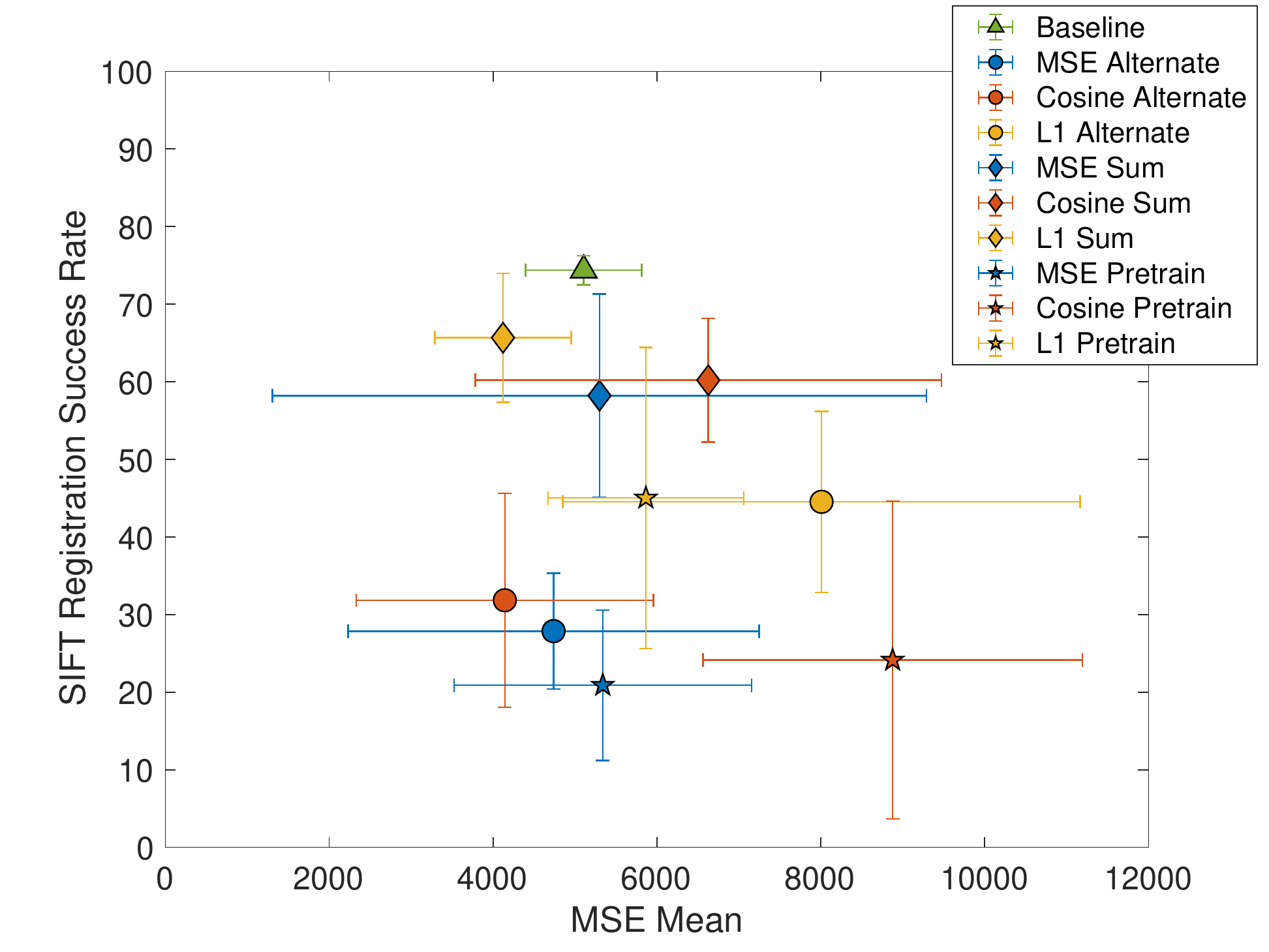}
         \caption{Reg. Succ. Rate vs MSE, \;$PCC = -0.24$}
         \label{fig:SIFTvsMSE}
     \end{subfigure}\hfill
     \begin{subfigure}[t]{0.32\textwidth}
         \centering
         \includegraphics[width=\textwidth,trim=13 8 5 0, clip]{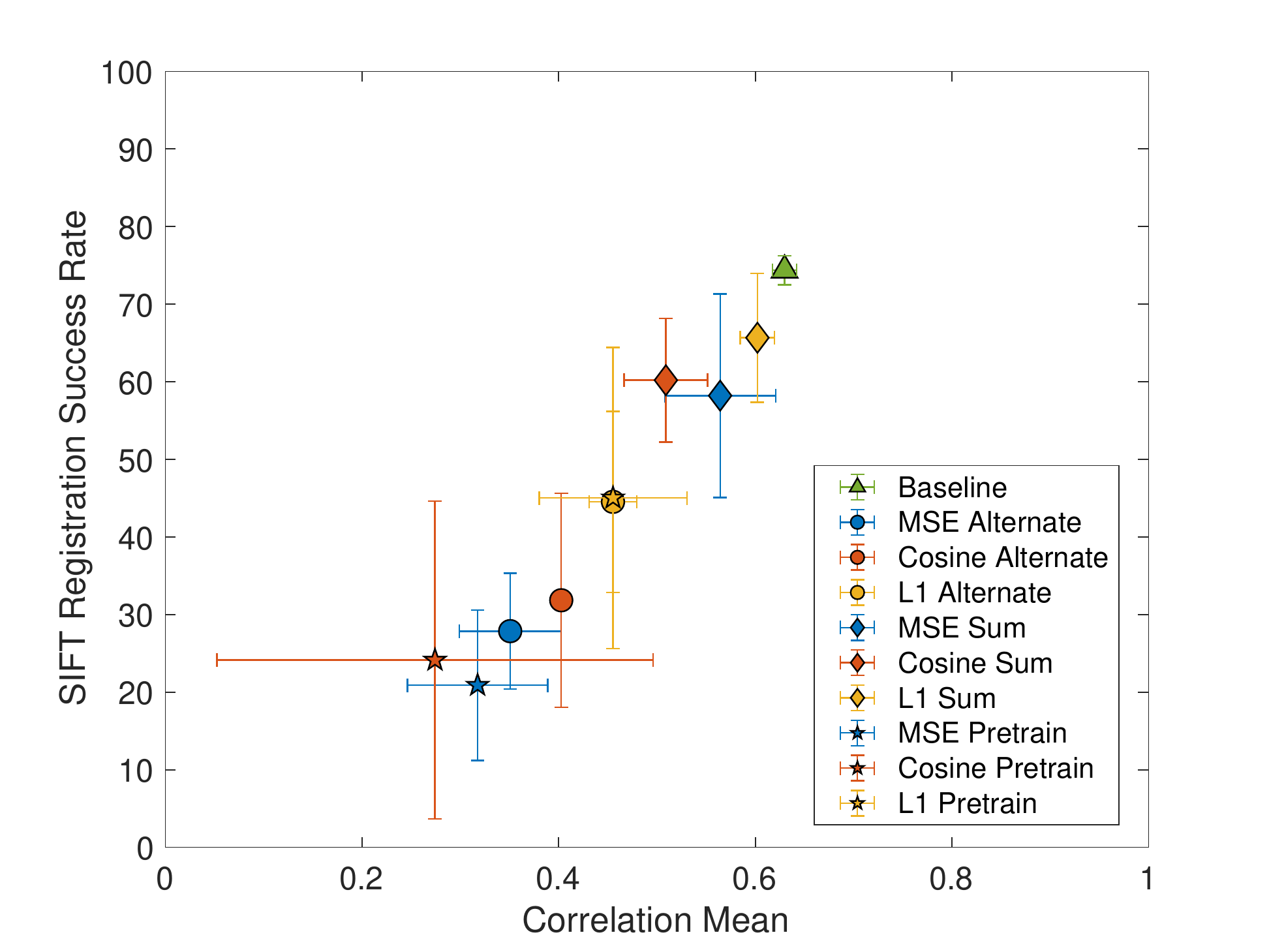}
         \caption{Reg. Succ. Rate vs Correlation, \;$PCC = 0.98$}
         \label{fig:SIFTvsCorr}
     \end{subfigure}\hfill
     \begin{subfigure}[t]{0.32\textwidth}
         \centering
         \includegraphics[width=\textwidth,trim=13 8 5 0, clip]{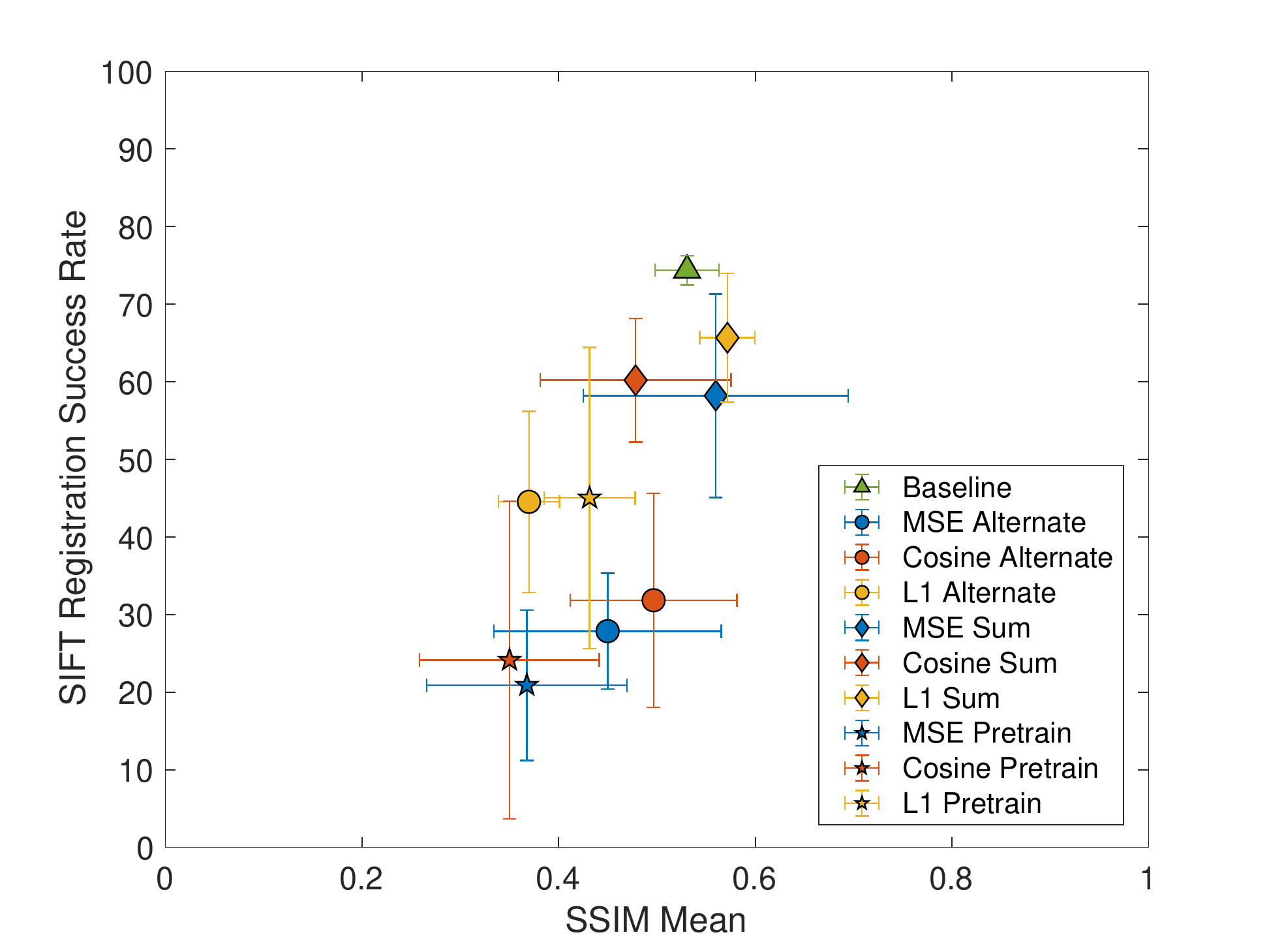}
         \caption{Reg. Succ. Rate vs SSIM, \;$PCC=0.75$}
         \label{fig:SIFTvsSSIM}
     \end{subfigure}
      \begin{subfigure}[t]{0.32\textwidth}
         \centering
         \includegraphics[width=\textwidth,trim=13 8 5 0, clip]{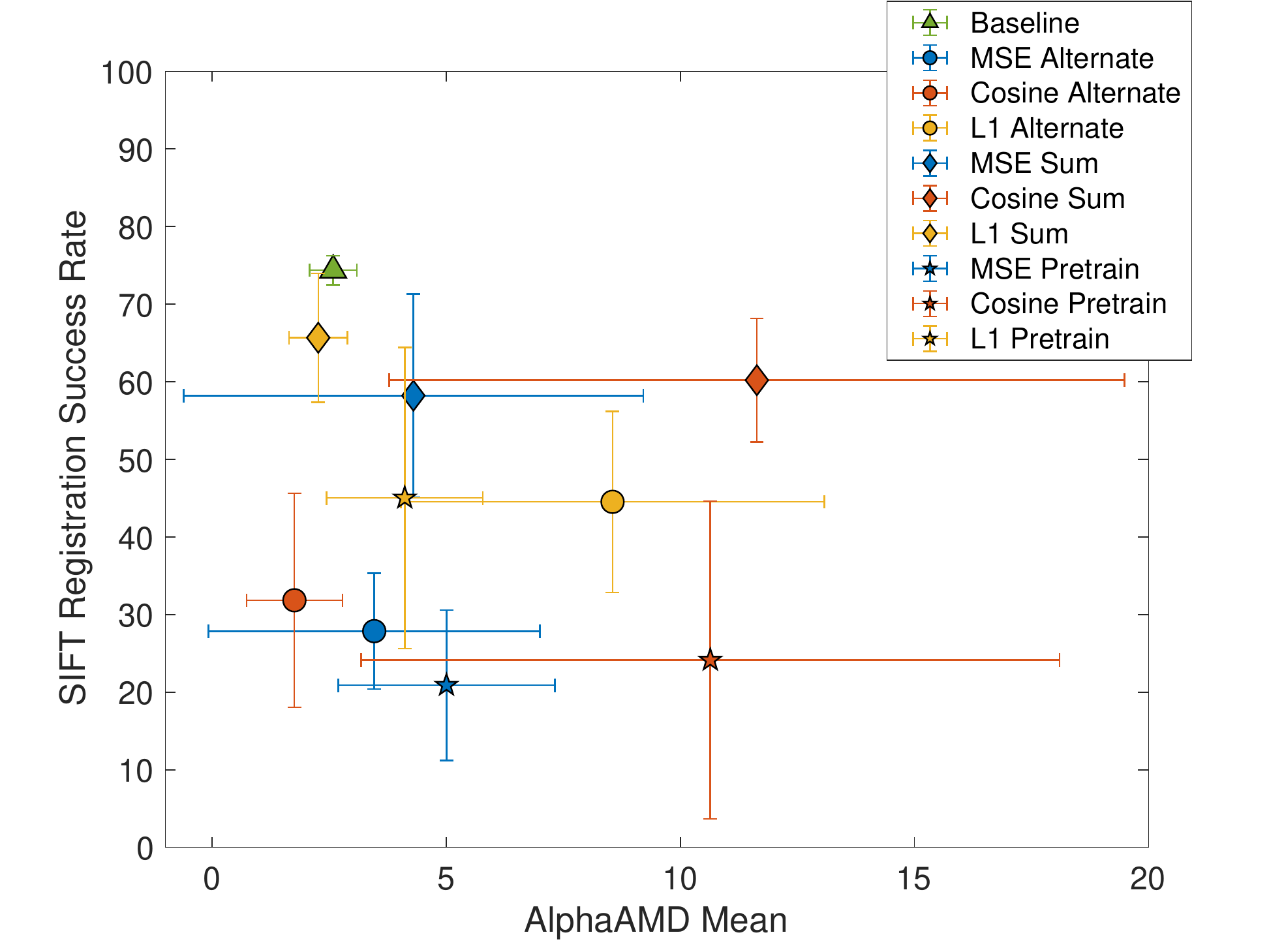}
         \caption{Reg. Succ. Rate vs \mbox{$\alpha$-AMD}, \;$PCC=-0.15$}
         \label{fig:SIFTvsAlphaAMD}
     \end{subfigure}\qquad
     \begin{subfigure}[t]{0.32\textwidth}
         \centering
         \includegraphics[width=\textwidth,trim=13 8 5 0, clip]{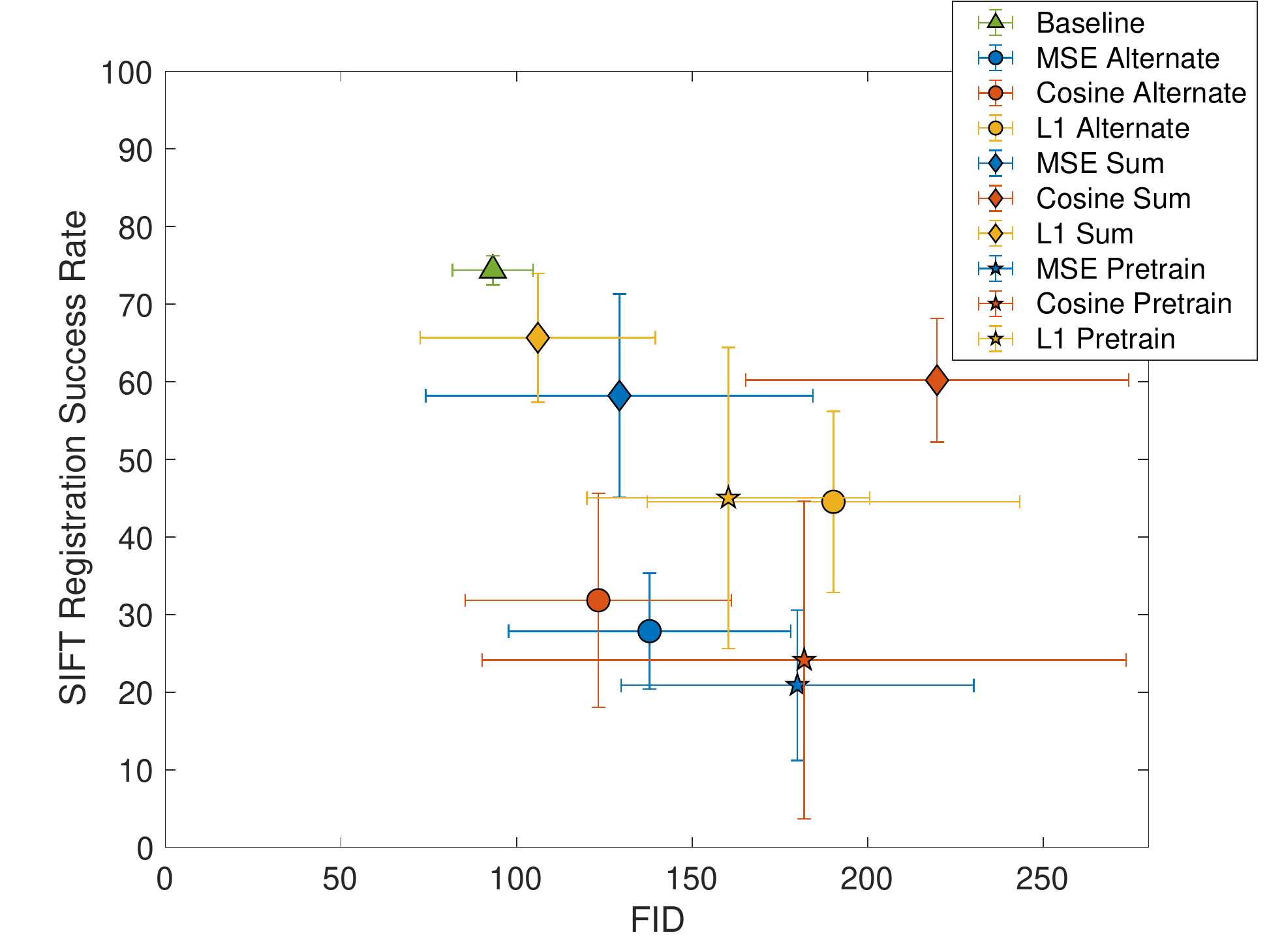}
         \caption{Reg. Succ. Rate vs FID, \;$PCC=-0.38$}
         \label{fig:SIFTvsFID}
     \end{subfigure}
        \caption{Relation between RSR 
        and image (dis-)similarity measures. Marker style indicates the approach (alternating, summed or pretraining loss), color of the marker and bars correspond to the 
        critic functions 
        used in $\mathcal{L}_{BN}(\mathcal{D})$ (none, MSE, cosine sim. and $L^1$), error-bars correspond to the standard deviations computed over 3 runs. The PCC between each measure and the registration performance is reported. Best appreciated zoomed in.}
        \label{fig:scatterplots}
\end{figure}
Fig.~\ref{fig:scatterplots} explores the relations between the mean of the considered quality assessment of the generated CoMIRs (in terms of their similarity/distance), and their RSR. 
Note that horizontal error-bars in Fig.~\ref{fig:scatterplots} can only be compared intraplot-wise, but not inter-plot, as the evaluation measures are of different ranges. The Pearson correlation coefficient (PCC) is reported to assess the linear relationship between the measures and RSR. The PCC ranges in $[-1,1]$ with $-1$ corresponding to a direct, negative correlation, $0$ representing no correlation, and $1$ representing a direct, positive correlation. Values close to ±1 indicate a high agreement between a similarity/distance measure and RSR.

\section{Exploration of the Embedding Space}
\label{sec:embeddings}
To investigate the reasons of the reduced performance when using additional supervision on the BN features, we inspect the training features on the BF \& SHG dataset. 

We can easily compute the relative (dis-)similarities between all CoMIR pairs, e.g. w.r.t. MSE. Based on dissimilarities of data points, Multidimensional Scaling (MDS, \cite{MDS}), used in dimensionality reduction and visualization, maps high dimensional data into a low dimensional metric space by numerical optimization, preserving the relative dissimilarities between the samples.

\begin{figure}
\centering
\includegraphics[width=\textwidth]{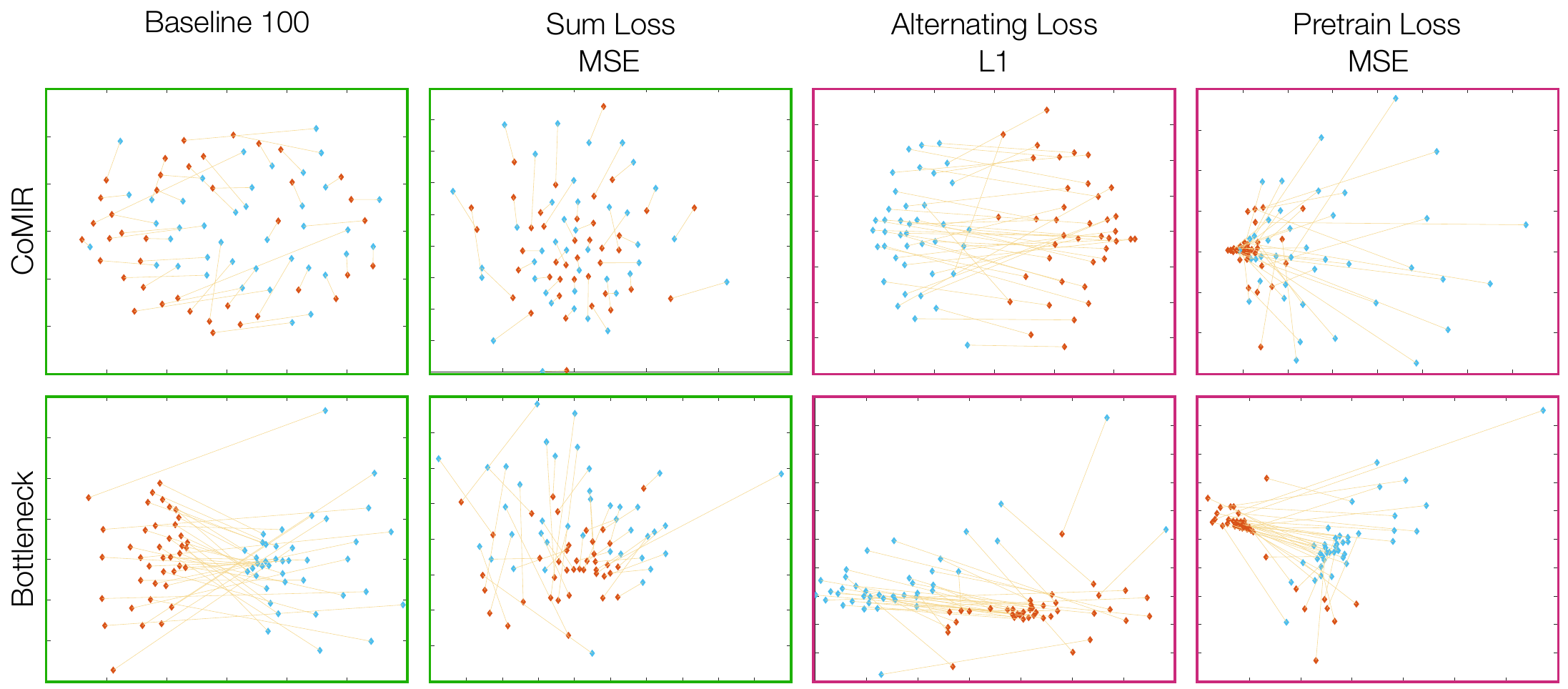}
\caption{Examples of Metric MDS embeddings of CoMIRs and BN features. BF features are marked by blue diamonds, SHG features by red diamonds, corresponding samples are connected by a yellow line. Green frames mark runs with high RSR (72.4\% for Baseline 100, 67.9\% for Sum Loss using MSE), the magenta frames low RSR (32.1\% for Alternating Loss using $L^1$, 9.7\% for Pretraining using MSE).}
\label{fig:MDSEmbeddings_main}
\end{figure}

We perform metric MDS on the dissimilarity matrix $\Delta=(d_{ij}) \in \mathbb{R}_+^{n \times n}$ to find 2D points whose distances $\bar{\Delta}=(\bar{d}_{ij})$ approximate the dissimilarities in $\Delta$. The resulting optimization problem minimizes an objective function called stress. We empirically observe that using Sammon's non-linear stress criterion~\cite{Sammon} results in the best fit and lowest stress to embed CoMIR and BN features, given by
\begin{equation}
\label{dissimilarity}
    \text{Stress}=\frac{1}{\sum_{i<j}d_{ij}}\sum_{i<j}\frac{(d_{ij}-\bar{d}_{ij})^2}{d_{ij}}
\end{equation}
where $d_{ij}$ denotes the distance between sample $i$ and $j$ in the high-dimensional space and $\bar{d}_{ij}$ the distance in the 2D projection space.
$\Delta$ contains the pairwise MSE between all features, either BN or CoMIR features.

Fig. \ref{fig:MDSEmbeddings_main} shows the MDS embeddings of BN and CoMIR features of the training set for selected runs of each approach. Visualizations for all runs are in appendix Sec. \ref{appendix:results}.
Features resulting from the BF images are marked by blue diamonds, from SHG images by red diamonds. All corresponding samples, i.e. the features resulting from a multimodal image pair, are connected by a yellow line. The green frames mark runs with high RSR, magenta frames runs with low RSR.
We observe in Fig. \ref{fig:MDSEmbeddings_main} that CoMIRs without intermediate loss on BN layers ("Baseline 100") are spatially spread. The distances between corresponding samples are reasonably small, though using a $\mathcal{L}_{BN}(\mathcal{D})$ with a MSE critic in a sum with $\mathcal{L}_{C}(\mathcal{D})$ can further reduce the distance between corresponding pairs. This embedding configuration is the best among all experiments w.r.t. pairwise distances, however the RSR is below the baseline.

\begin{wrapfigure}{r}{0.5\textwidth}
  \centering
  \vspace{-22pt}
    \includegraphics[width=0.48\textwidth,trim=0 2 0 0,clip]{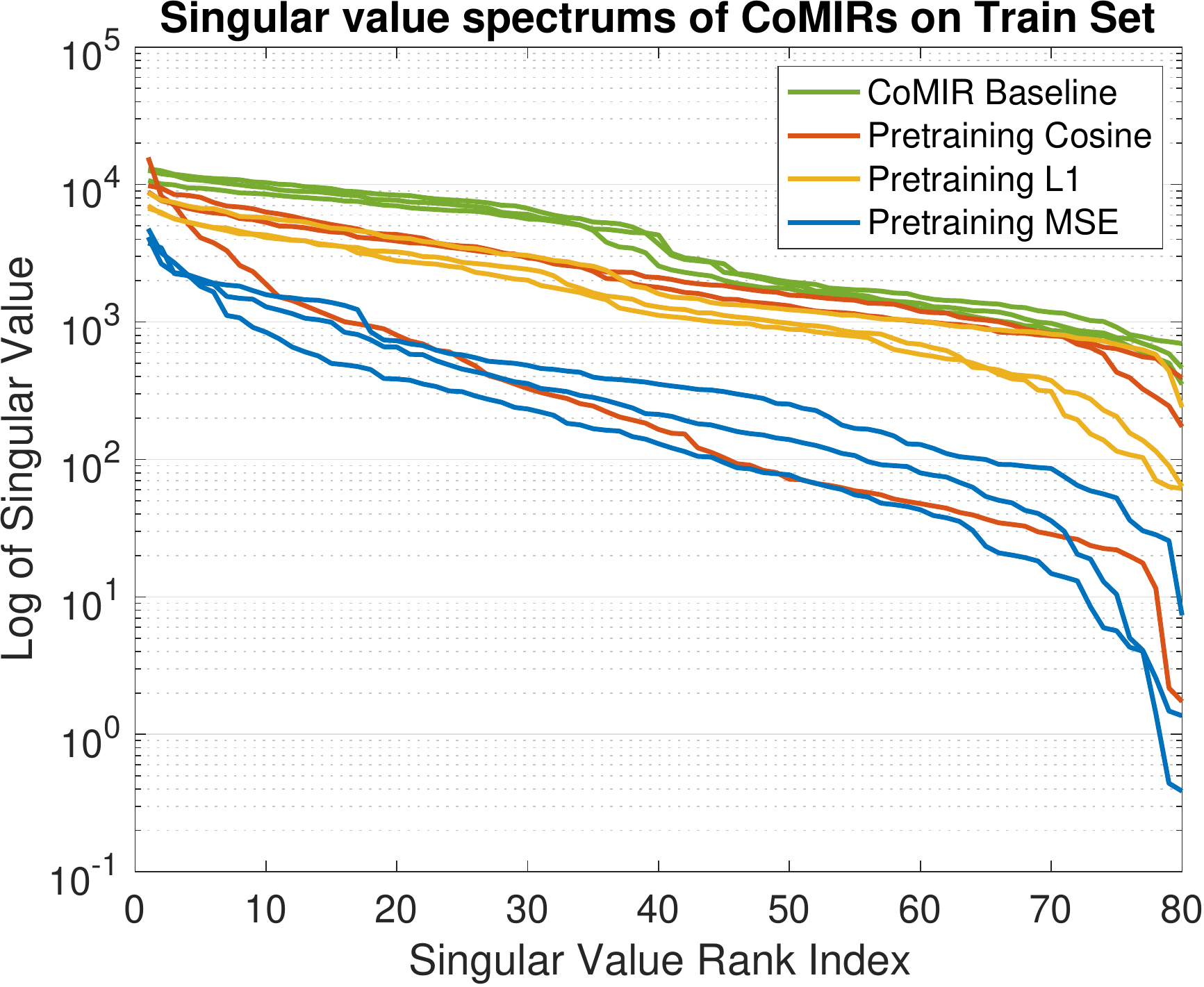}
  \caption{Log-Plot of the SVs of CoMIRs learned after pretraining with a loss on the BN, and the baseline. Three runs were performed for each experiment.}
  \label{fig:SVD}
    \vspace{-22pt}
\end{wrapfigure}

For some experiments with low RSR, we observe that the embeddings have a tendency to cluster w.r.t. the modalities, as for example using an alternating loss with $L^1$ in Fig. \ref{fig:MDSEmbeddings_main}. Furthermore, we observe that the pretraining approach, can lead to a cluster of SHG samples, contracted to a single point.
Additionally, we visualize the embeddings of the BN features for each of these approaches (visualizations for all runs are in the appendix). We observe that training CoMIRs without intermediate supervision results in BN features which are clustered by the modality they originate from. 

 
We observe that CoMIRs of SHG images learned after pretraining the BN layer can collapse to a point. To quantify the dimensionality of the embedding space - or its collapse - the authors in \cite{Jing2021UnderstandingDC} compute the covariance matrix of the embedding vectors and perform its singular value decomposition (SVD). Inspecting the spectrum of the resulting singular values (SVs) by plotting their value in sorted order, shows how many collapse to zero, corresponding to collapsed dimensions. This approach to quantify DC in CL has also been adopted in \cite{https://doi.org/10.48550/arxiv.2205.06926,10.1007/978-3-031-19821-2_28}. \\
We plot the SV spectrum of the covariance matrix of CoMIRs in Fig. \ref{fig:SVD}. While we do not observe a complete DC, we see that the SVs decay much faster for approach III using pretraining, i.e. they are closer to a low-rank solution. The square of a SV is proportional to the variance explained by its corresponding singular vector, i.e. the smaller the SVs the less variance in the data is explained and more likely corresponds to noise.

\section{Discussion}
Our results shown in Fig.~\ref{fig:SIFTBarplot} and \ref{fig:CytoBarplot} indicate that CoMIRs generated using a loss acting only on the final features yield the best representations for the downstream task of registration on both observed datasets, outperforming all evaluated attempts to impose similarity of the features earlier in the networks.
We further observe
on both datasets that Approach II leads to a higher RSR than Approach I and III. 
On the BF \& SHG Dataset, using $L^1$ as a similarity in the contrastive loss produces, on average, better representations w.r.t.\ RSR, compared to cosine similarity, which in turn performs better than using MSE. However, the error bars indicate a wide spread in the results.
On the QPI \& Fluorescence Dataset, Approach III results in CoMIRs with higher RSR compared to Approach I (alternating the losses), unlike in the case of BF \& SHG images.



Figure~\ref{fig:scatterplots} presents the relation between similarity/distance of generated CoMIR pairs and their RSR. This connects our two stated evaluation objectives: the quality of appearance of the representations, and their applicability for registration. 
For the distance measures (MSE, $\alpha$-AMD, FID) a well performing measure would be located in the upper left corner (high RSR, and low distance between the image pairs), while for the similarity measures (Correlation, SSIM) it would be in the upper right corner. 
We observe that the mean correlation between CoMIR pairs (Fig.~\ref{fig:SIFTvsCorr}) correlates strongly with the RSR of these representations 
(PCC 0.98), while for the other measures the relations are less clear. As intuitively expected, the PCC indicates (weak) negative correlation between RSR and distance measures and positive correlation for the similarity measures.
We observe that for the BF \& SHG Dataset the baseline approach performs best in terms of RSR, also exhibiting stability (low standard deviation), high similarity and low distance between the corresponding image pairs.
Among the experiments using intermediate losses on the BN, $L^1$-based loss generated the most correlated CoMIRs within any loss fusion approach and the highest RSR.

Furthermore, we investigate the reasons of the lower RSR by visualizing the embeddings of the train set in Sec. \ref{sec:embeddings}. We observe that additional supervision on the BN can lead to clustering by modality in the embedding space, and in turn to less discriminative features among the corresponding CoMIR pair. The embedding of the BN features shows that the network learns features at the BN which are modality-specific rather than similar across modalities when left unregulated. This is in line with other observations that suggest pseudo-Siammese networks are prefered over Siammese networks for multimodal tasks in which the modalities differ strongly from each other~\cite{8451804}. 

We observe that pretraining on the BN features can lead to partial DC from which the training cannot recover in the subsequent training of the final layer. This observation may be related to observations in \cite{10.1007/978-3-031-19821-2_28}, which connect partial DC to models which are too small relative to the dataset size. However, this was observed for SimSiam, a non-contrastive SSL approach not relying on negatives.

\section{Conclusions}
Contrastive learning can generate common representations for multimodal images, called CoMIRs, which are similar in intensity, structures and features, allowing the utilization of monomodal registration methods. This reduces the very challenging task of multimodal registration to a, typically easier, monomodal one. In this study, we explore three approaches to add supervision to the latent representations in the bottleneck of the U-Net used to generate CoMIRs, imposing similarity of the features earlier in the network. For each approach we test three critic functions for the loss evaluated on these latent features. Our results show that CoMIRs learned with no additional supervision perform best in the downstream task of registration. More so, we show that without additional supervision, the BN features tend to extract modality-specific information and the shared features are likely extracted in the decoder of the U-Net. Additional supervision on the BN features often propagates the tendency to modality-wise clusters in the feature space into the final CoMIRs, making them less useful for registration. We observe partial DC of the learned features when the contrastive loss is only applied to the BN during pretraining and see that this collapse is maintained during the subsequent training on the final layers.


We address the quality quantification of learned image representations. We relate commonly used image distance and similarity measures to the representations' usefulness for registration. We show that correlation corresponds highly to registration performance. Representations useful for registrations also score reasonably well with respect to SSIM, MSE, $\alpha$-AMD, and FID, however the reverse does not necessarily hold.
The study explores the behavior of CoMIRs with BN supervision on a publicly available multimodal dataset of BF and SHG images. 
We confirm the main finding that CoMIRs without additional intermediate supervision are more useful for registrationon on a second multimodal dataset of correlative time-lapse QPI and fluorescence images.

Our study indicates the importance to develop representation learning approaches with a particular application on mind, as we show that concepts applicable for biomedical image classification do not necessarily generalize to registration tasks in the same domain and learning context, in this case contrastive learning. Multimodal image registration remains in general a very difficult task and we believe it is important to explore and further improve upon suitable learning strategies, which we will continue in our future work.
%
%
\bibliographystyle{splncs04}
\bibliography{main}
\clearpage
\section{Appendix}

\subsection{Performance measures}
\label{appendix:measures}
Finding a suitable measure to assess the quality of representations for the downstream task of registration is challenging. We evaluate the quality of the differently learned dense, image-like CoMIRs in two ways, both through their direct comparison using several image similarity/distance measures, and by evaluation of their performance in the downstream task of image registration. 
The measures used in this studies are defined as follows.

%
\noindent
\textbf{MSE}\\
The MSE is computed by averaging the squared pixel-wise intensity differences between two images. It is given by 
$$ MSE(Y^A, Y^B)= \frac{1}{N} \sum_{i=1}^N ||Y_i^{A}-Y_i^{B}||_2$$ for $Y_i$ denoting pixels of aligned corresponding CoMIRs of  modality A and B, respectively. $N$ denotes the total number of pixels in each image.
A lower MSE value indicates higher similarity between two images (CoMIRs).\\

\noindent
\textbf{SSIM}\\
The SSIM index \cite{1284395} assesses the visual impact of luminance, contrast and structure. It is calculated using a sliding Gaussian window over the image pair. The local SSIM is given by

$$
SSIM(Y_w^A, Y_w^B) = \frac{(2\bar{Y}_w^A\bar{Y}_w^B+c_1)(2\sigma_{Y_w^AY_w^B}+c_2)}{\left((\bar{Y}_w^A)^2(\bar{Y}_w^B)^2+c_1)(\sigma_{Y_w^A}^2+\sigma_{Y_w^B}^2+c_2)\right)} 
$$
where $Y_w$ is a window in one modality of chosen size to compute the local SSIM, $\bar{Y}_w$ denotes its mean, $\sigma_{Y_w}$ its standard deviation, $\sigma_{Y_w^AY_w^B}$ the cross-covariance between two windows of different modalities, $c_1 = (0.01 \cdot L)^2$, $c_2 = (0.03 \cdot L)^2$, where $L$ is the dynamic range value. The global SSIM is the mean of local SSIM values and ranges in $[0, 1]$. The value 1 indicates that the two images (CoMIRs) have maximal structural similarity. Smaller values indicate deviations.\\

\noindent
\textbf{Correlation}\\
The 2-D correlation coefficient between two aligned images (CoMIRs) generated from two input modalities is given by
$$r(Y^A, Y^B)=\frac{\sum_{m} \sum_{n} (Y_{mn}^A-\bar{Y}^A)(Y_{mn}^B-\bar{Y}^B)}{\sqrt{\left(\sum_{m} \sum_{n}(Y_{mn}^A-\bar{Y}^A)^2\right)\left(\sum_{m}\sum_{n}(Y_{mn}^B-\bar{Y}^B)^2\right)}}$$
where $\bar{Y}$ is the mean of the respective images of the respective modalities.\\

\noindent
\textbf{$\alpha$-AMD}\\
Distance measures between images combining intensity and spatial information, were introduced in \cite{8643403} and further developed for intensity-based registration in \cite{6645422}. The developed registration framework is successfully combined with CoMIRs for multimodal registration in \cite{pielawski2020comir}. Here we use $\alpha$-AMD to quantify the distance between two aligned CoMIRs generated by the two input modalities. Further details about the computation, as well as the related code are available at \cite{johancode}.\\

\noindent
\textbf{FID}\\
FID \cite{NIPS2017_8a1d6947} has been used as a distance measure between two sets of images to quantify the quality of GAN-generated images. It has been shown to correlate well with human visual assessments of generated images \cite{drti,9157662,DBLP:journals/corr/abs-2101-08629}; however, counter examples are given in \cite{https://doi.org/10.48550/arxiv.1803.07474}. Correlation between FID as a measure of quality of generated CoMIRs and registration performance has been reported in \cite{10.1371/journal.pone.0276196}. Unlike the other measures listed above, FID is not being computed for one image pair, but evaluates the similarity between the distributions of two sets of images. It is computed using the output activations of an Inception v3 \cite{7780677} network, pretrained on ImageNet \cite{5206848} and considers the distribution of all activations of the test set. It is given by
$$
FID(\mathcal{D}^A, \mathcal{D}^B) = || \mu^A - \mu^B ||_2^2 + tr\left( \Sigma^A + \Sigma^B - 2 \left( (\Sigma^A)^{\frac{1}{2}} \cdot \Sigma^B \cdot (\Sigma^A)^{\frac{1}{2}} \right)^{\frac{1}{2}} \right)
$$\\
where $\mu$ is the mean vector and $\Sigma$ the covariance matrix of a multivariate distribution resulting form a set of images in its respective modality.

\subsection{Implementation Details} 
\label{appendix:implementation}
The U-Net architectures \cite{jegou2017one} for both the BF \& SHG dataset and QPI \& Fluorescence dataset are identical. They have 32 convolutional filters for the first convolution, 4 dense blocks of depth 6 as down and up blocks and 4 BN layers. Upsampling was used as well as max pooling, a dropout rate of $0.2$, no early transition or activation function in the last layer and no compression. There is no non-linear activation in the final layer.

The temperature in the loss on the final layer was set to $0.5$ for all experiments.

Stochastic gradient descent was used with a learning rate of $10^{-2}$, a weight decay of $10^{-5}$ and a momentum of $0.9$. The batch size was $32$, and the steps per epoch $32$. The gradient norm was limited to $1$. $L_1$ activation decay was set to $10^{-4}$, $L_2$ activation decay to $10^{-2}$. The image patch size was $128 \times 128$. The data augmentation consisted of flips ($p=0.5$) and random integer rotations by up to $\pm 180\circ$ using either a linear, nearest neighbor or cubic interpolation randomly, Gaussian blur ($p=0.5, \sigma=(0, 2.0)$), additive Gaussian noise(loc=0.0, scale=(0.0, 0.1)) and linear contrast adjustment ($p=0.5$, scaling (0.8, 1.0/0.8)).

For approach I, the hyperparameter to combine the losses was set to 1. For approach II, the hyperparameter was $0.5$.

The registration using SIFT is based on the implementation in Fiji 2.0.0 using the \verb+mpicbg.imagefeatures+ package. The feature descriptor size is 4 samples per row and column, the orientation bins are 8 bins per local histogram. The scale octaves are in $[128,1024]$px with 3 steps per scale octave and the initial $\sigma$ of each scale octave is equal to $1.6$.

\subsection{Results}
\label{appendix:results}
\begin{figure}[tb]
    \centering
    \begin{subfigure}[t]{0.32\textwidth}
        \includegraphics[width=\textwidth,trim=20 0 43 0, clip]{figures/barplot.pdf}
         \caption{RSR in percent.}
         \label{fig:SIFTbarplot_appendix}
     \end{subfigure}\hfill
    \begin{subfigure}[t]{0.32\textwidth}
        \includegraphics[width=\textwidth,trim=11 0 52 0, clip]{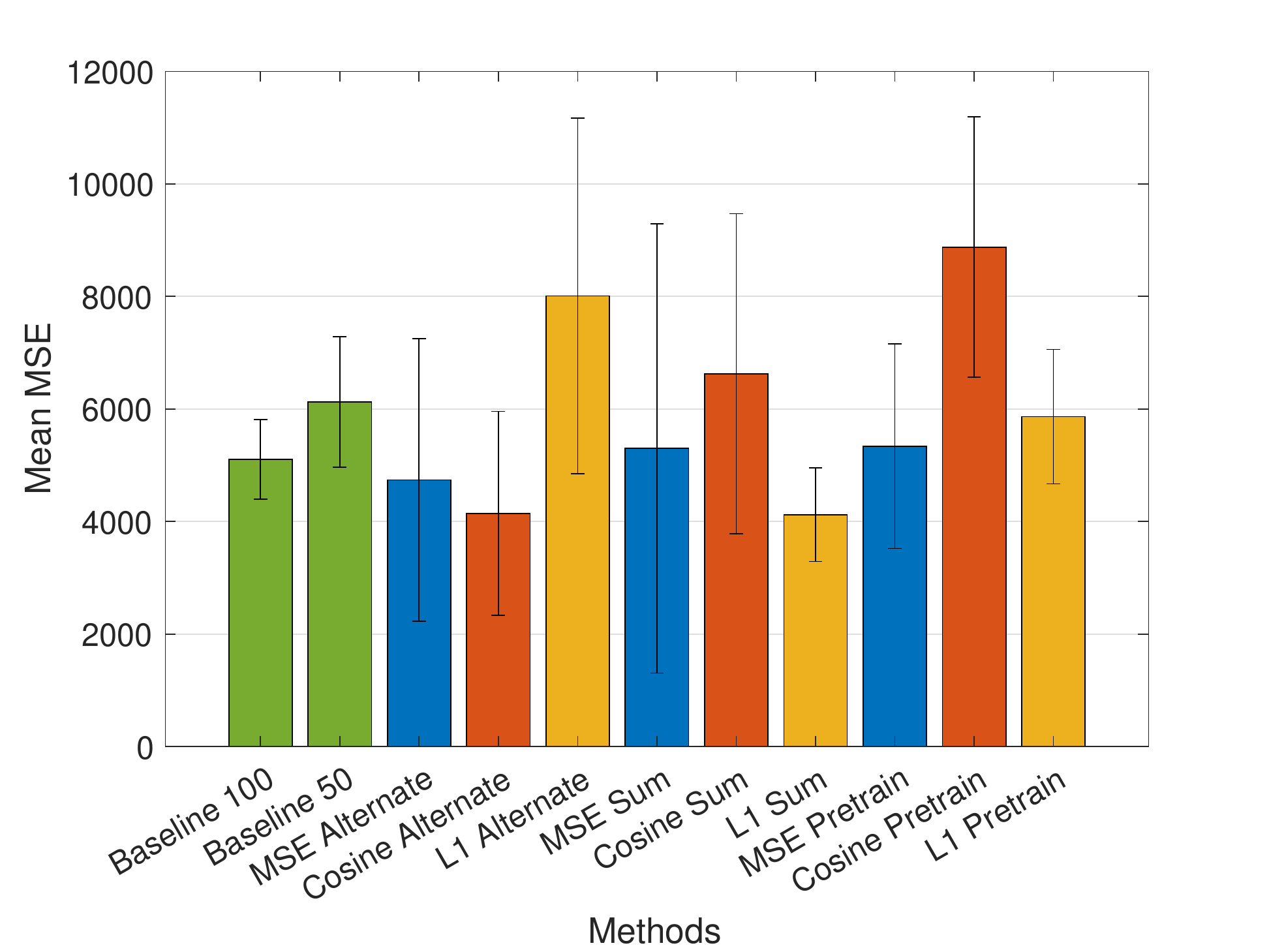}
         \caption{MSE}
         \label{fig:MSEbar}
     \end{subfigure}\hfill
     \begin{subfigure}[t]{0.32\textwidth}
         \includegraphics[width=\textwidth,trim=20 0 43 0, clip]{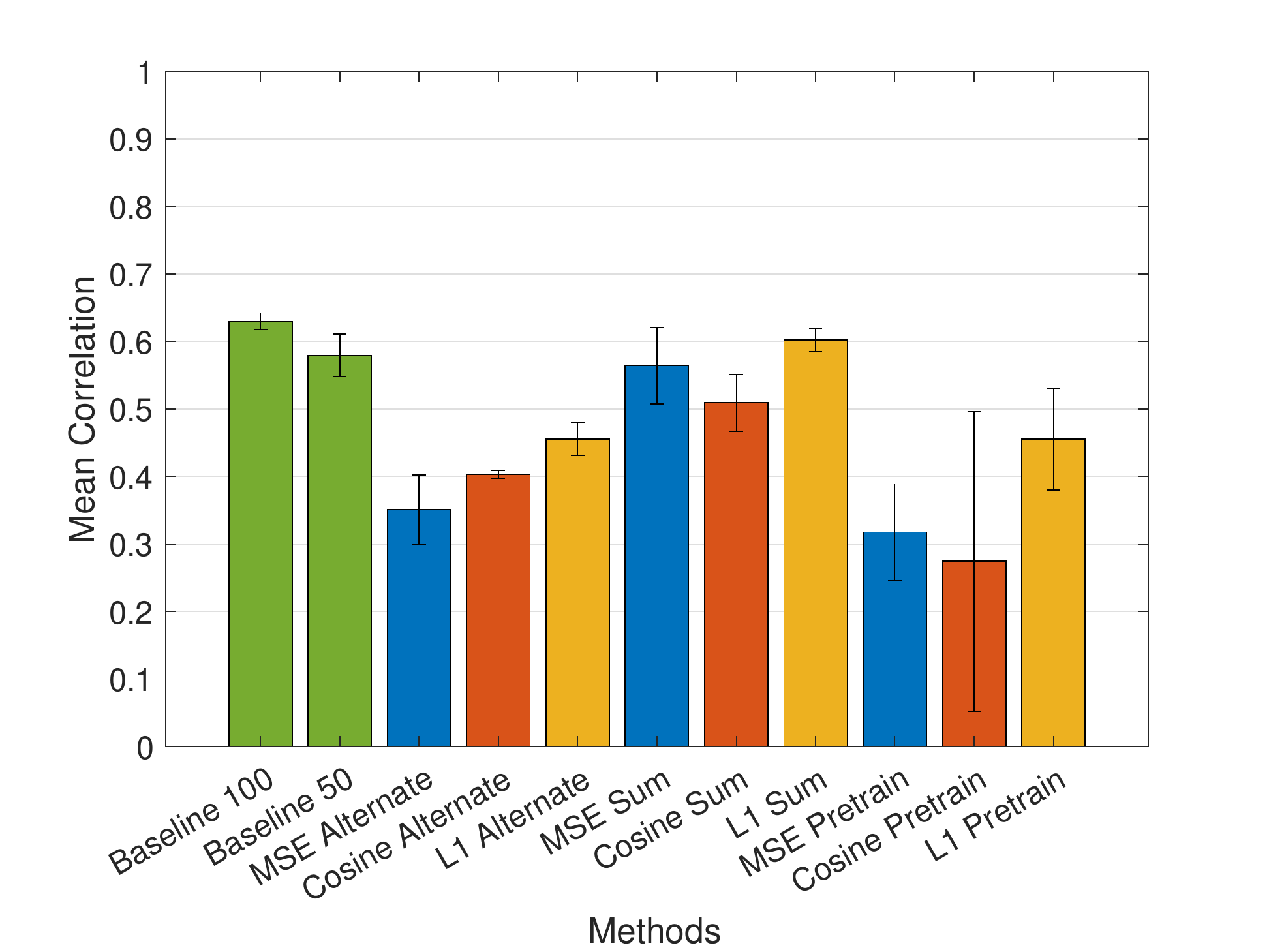}
         \caption{Correlation}
         \label{fig:Corrbar}
     \end{subfigure}
     \begin{subfigure}[t]{0.32\textwidth}
         \includegraphics[width=\textwidth,trim=20 0 43 0, clip]{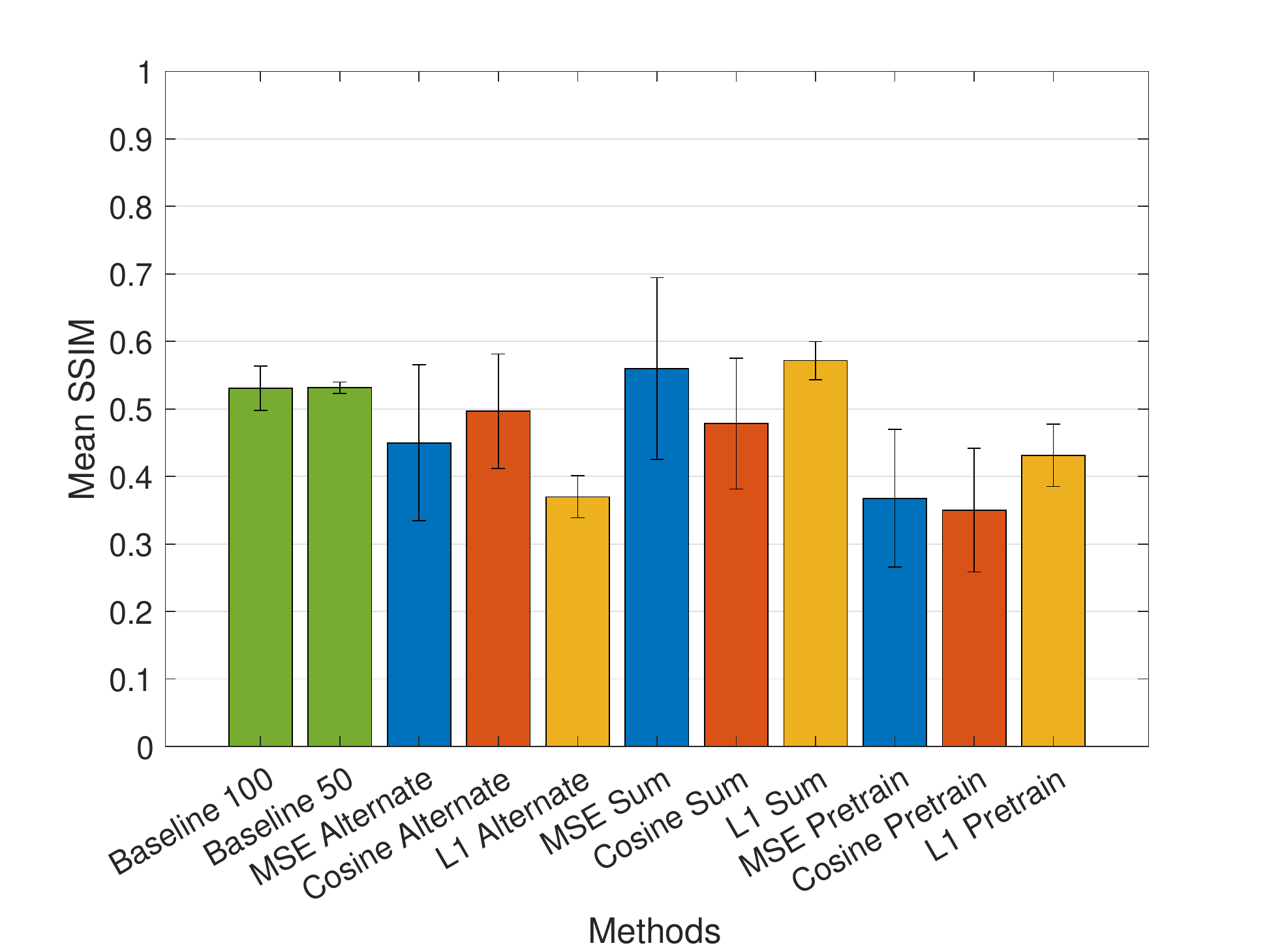}
         \caption{SSIM}
         \label{fig:SSIMbar}
     \end{subfigure}\hfill
    \begin{subfigure}[t]{0.32\textwidth}
         \includegraphics[width=\textwidth,trim=20 0 43 0, clip]{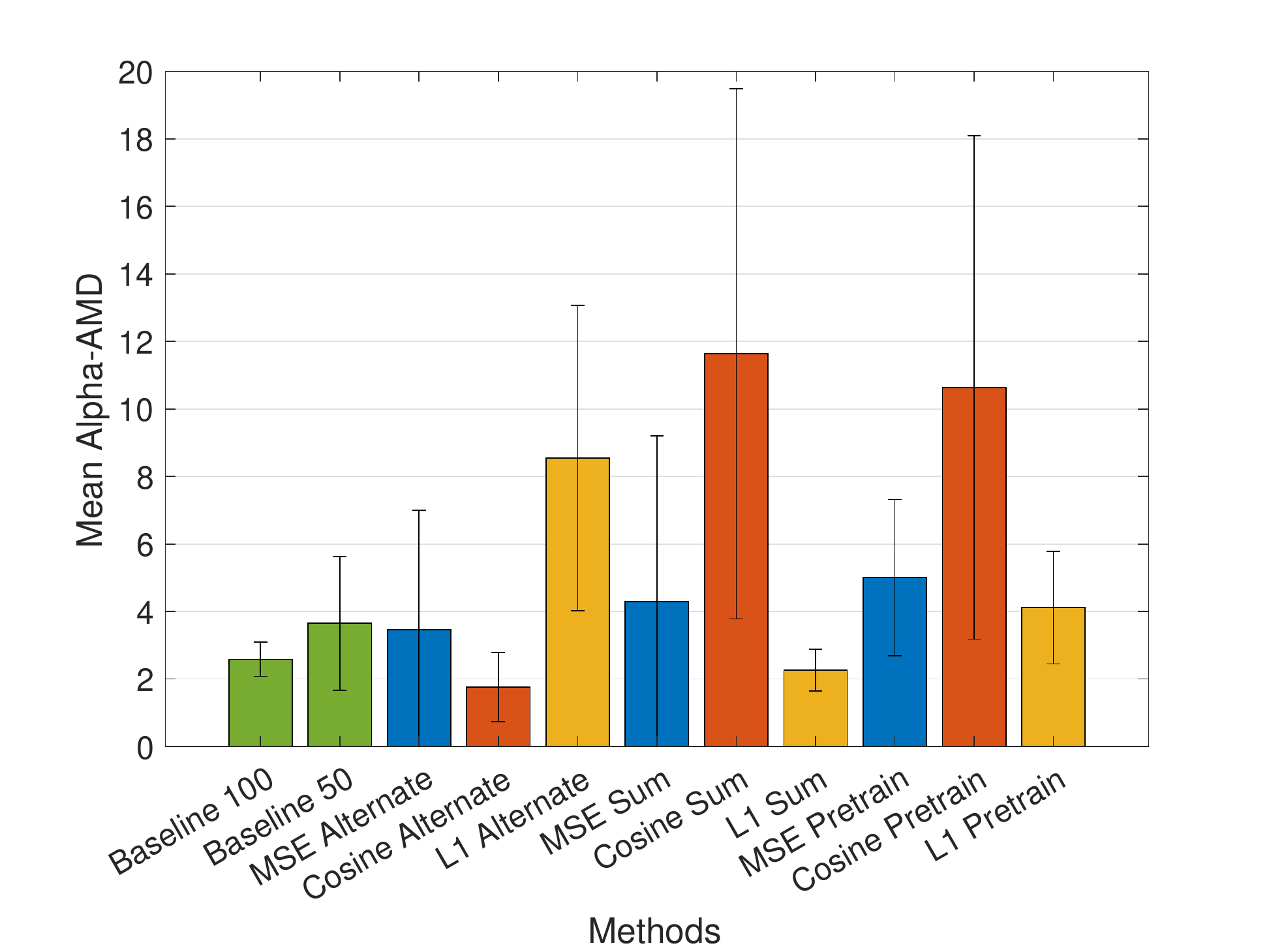}
         \caption{Alpha-AMD}
         \label{fig:AlphaAMDbar}
     \end{subfigure}\hfill
     \begin{subfigure}[t]{0.32\textwidth}
         \includegraphics[width=\textwidth,trim=20 0 43 0, clip]{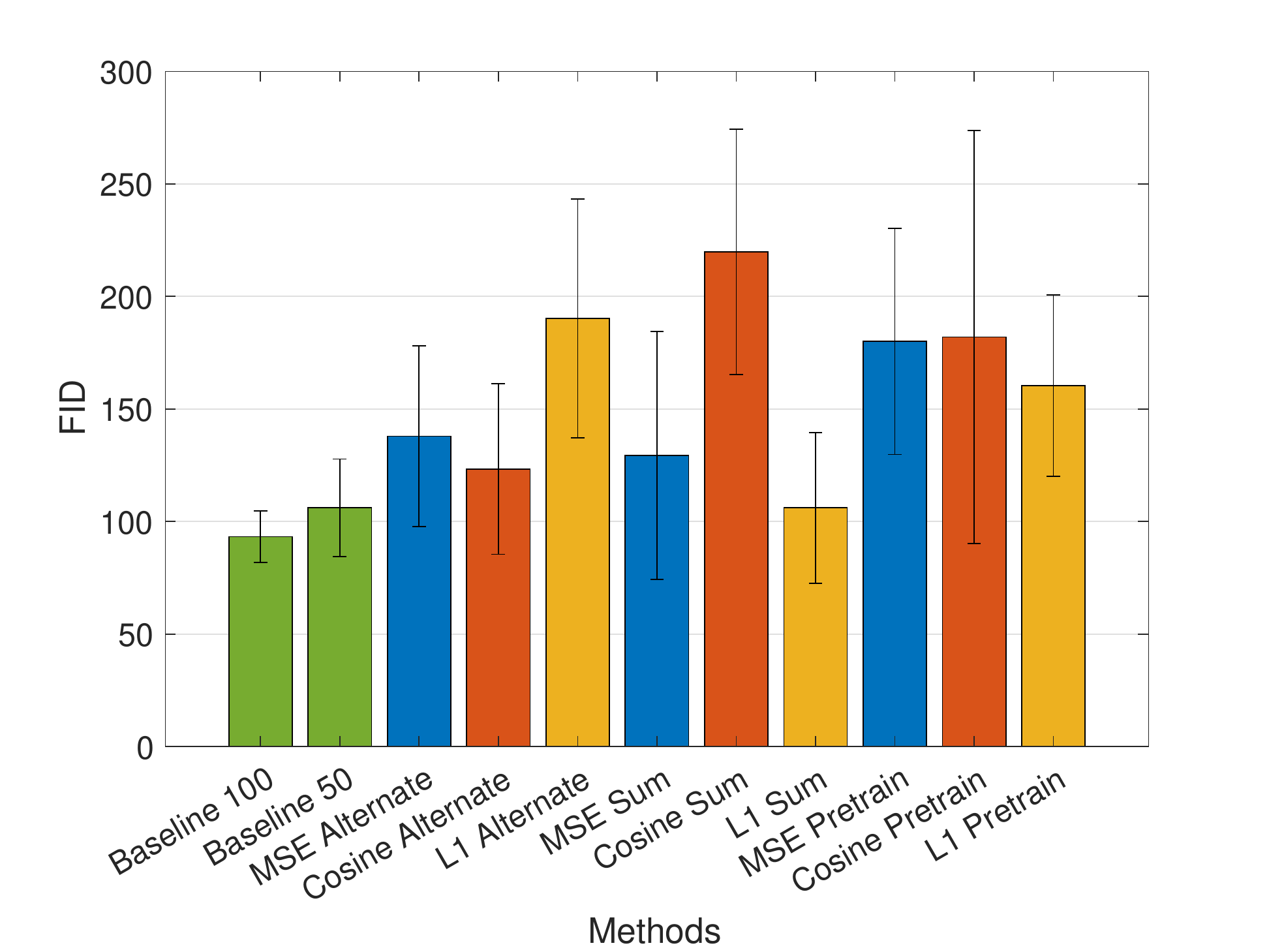}
         \caption{FID}
         \label{fig:FIDbar}
     \end{subfigure}
        \caption{Evaluation of the generated CoMIR representations by different approaches of supervision of the BN latent representations. (a) Success rate of registration of the generated  CoMIRs, for all the considered representation learning methods. (b)-(f) Quality of the generated CoMIRs, expressed by different similarity (c,d), or distance (b,e,f) measures, for all the considered representation learning approaches. "Baseline 100" and  "Baseline 50" refer to the baseline trained for 100 and 50 epochs, respectively. The color of the bars corresponds to the similarity/distance measure used in the loss for the BN features (green-none, blue-MSE, red-cosine similarity and yellow-$L^1$-norm). The error-bars correspond to the standard deviation computed over 3 runs.}
        \label{fig:barplots}
\end{figure}

Detailed results of experiments with the BF \& SHG dataset are given in Table \ref{tab:histo_all} and for the QPI \& Fluorescence dataset in Table \ref{tab:cyto_all}.

\begin{table}
\includegraphics[width=\textwidth]{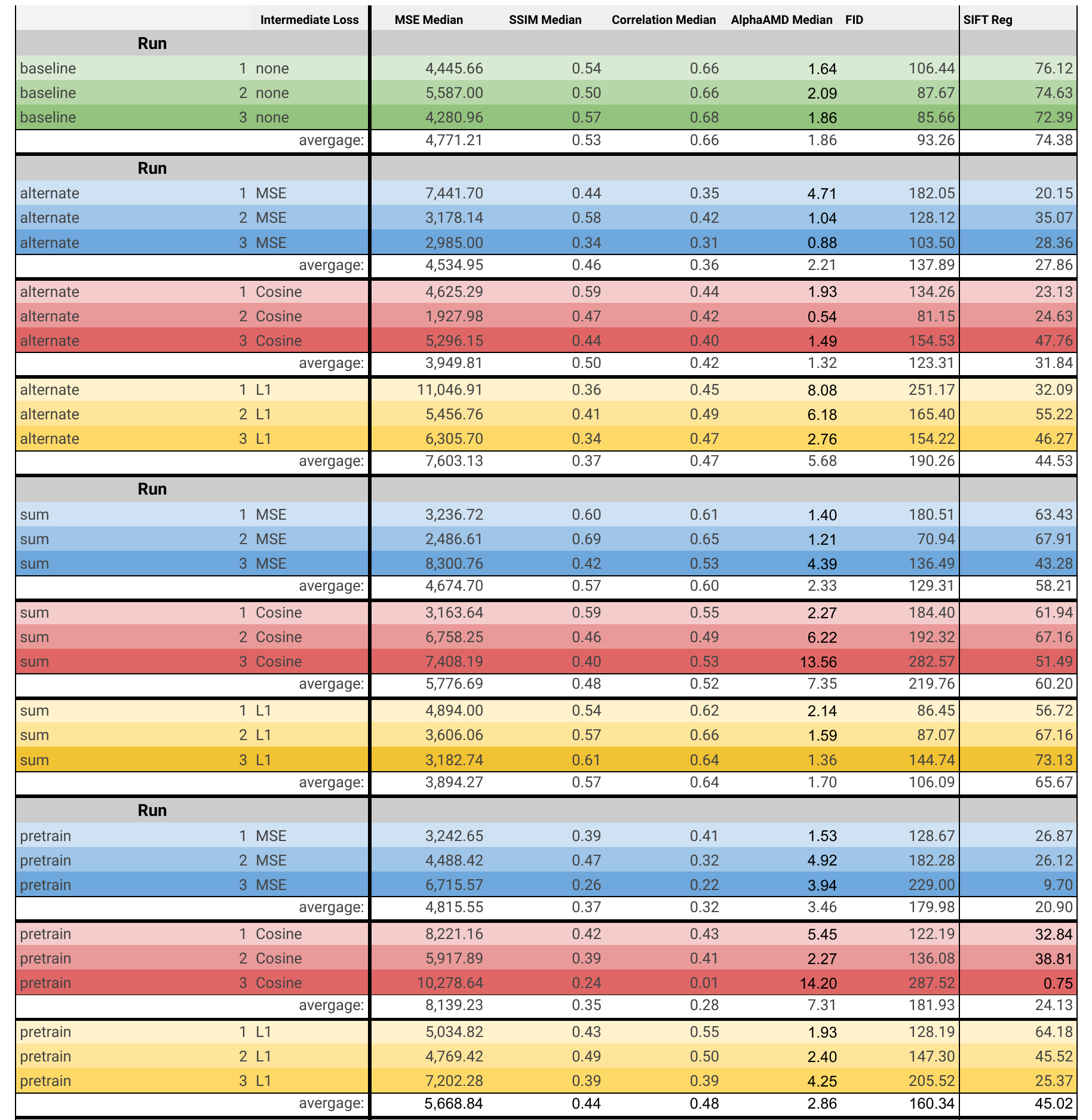}
\caption{Registration Results and image metrics on BF \& SHG dataset.}
\label{tab:histo_all}
\end{table}

\begin{table}
\centering
\includegraphics[height=0.9\textheight]{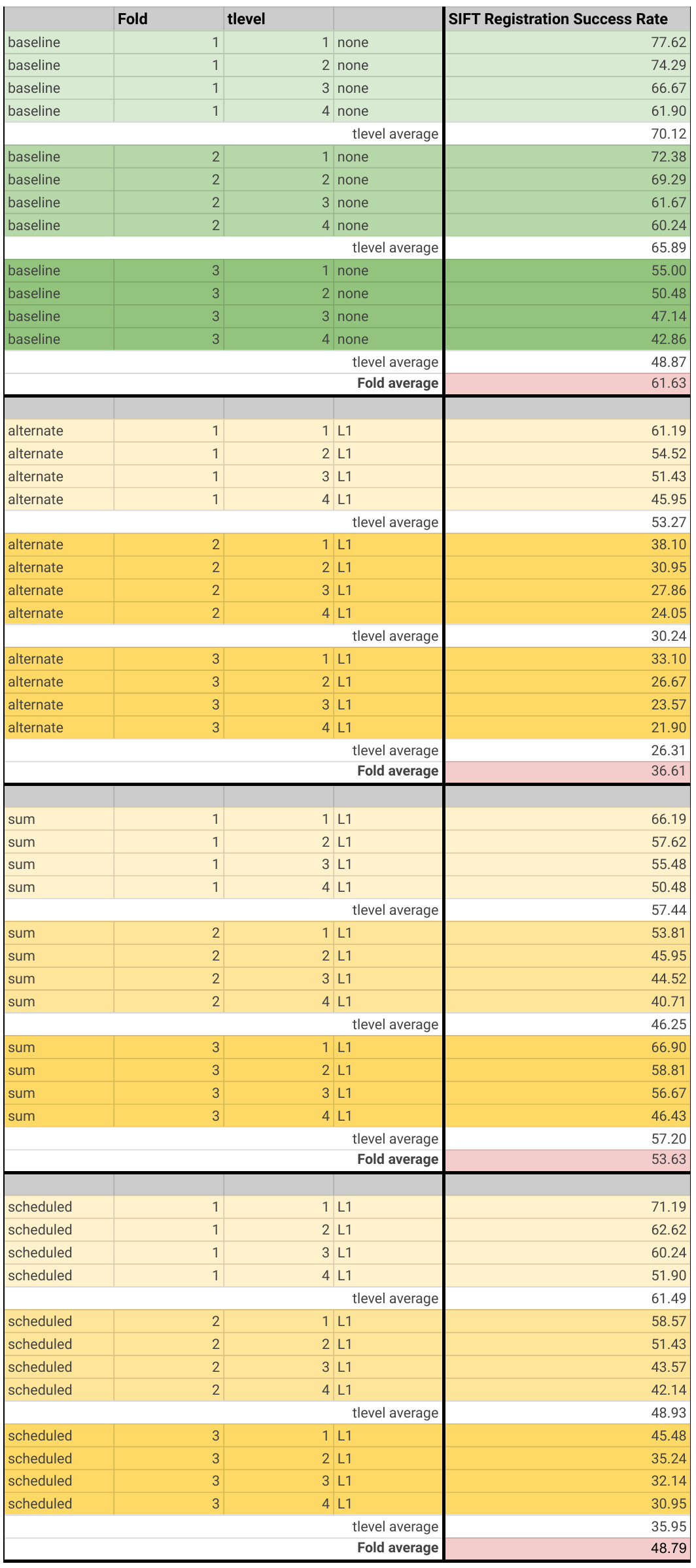}
\caption{Registration Results on Fluorescence \& QPI dataset.}
\label{tab:cyto_all}
\end{table}

\begin{figure}[tb]
     \centering
     \begin{subfigure}[b]{1\textwidth}
         \centering
         \includegraphics[width=\textwidth]{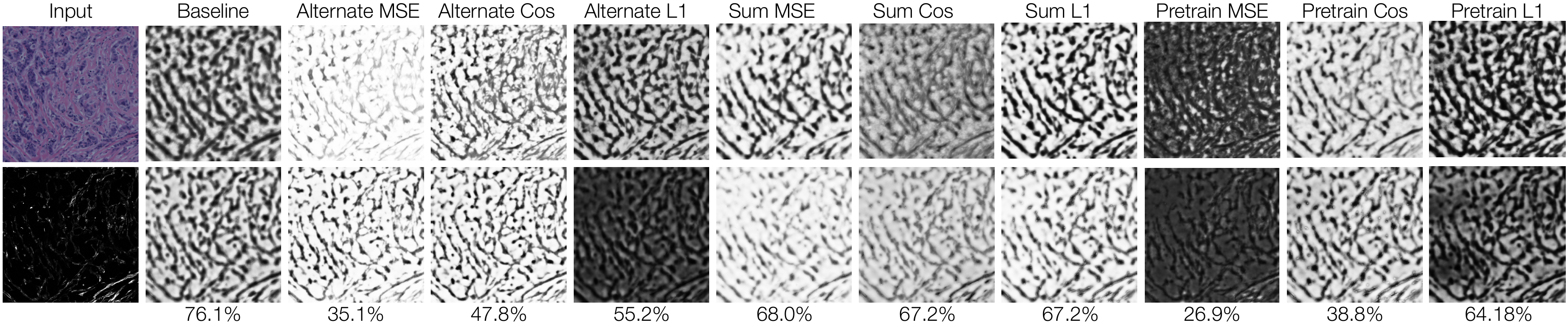}
         \caption{CoMIRs of runs with highest registration performance for each approach.}
         \label{fig:good}
     \end{subfigure}
     
     \begin{subfigure}[b]{1\textwidth}
         \centering
         \includegraphics[width=\textwidth]{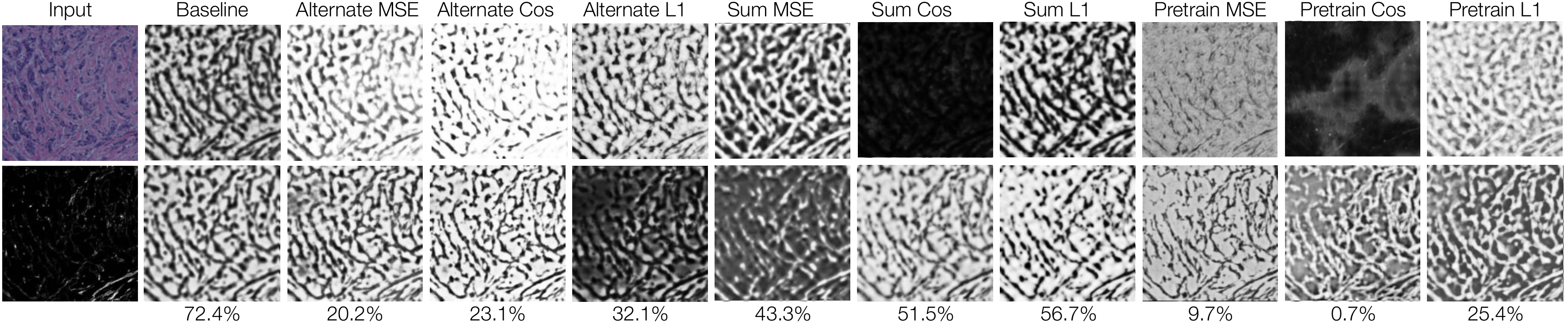}
         \caption{CoMIRs of runs with lowest registration performance for each approach.}
         \label{fig:bad}
     \end{subfigure}
            \caption{Example of one image pair from the BF \& SHG test set and its CoMIRs produced by the studied approaches. (a) The representations generated by the best performing run w.r.t. to the RSR of one particular setup for that image pair.  (b) The representations generated by the worst performing run of one particular setup for that image pair. The upper rows in (a) and (b) show the BF image and its corresponding CoMIRs, the lower rows in (a) and (b) show the SHG image and its CoMIRs. Below the CoMIRs, the RSR on the entire test set, reached in the particular run that generates the shown CoMIRs, is given for reference.}
        \label{fig:reps}
\end{figure}

Figure~\ref{fig:reps} shows an image pair randomly selected from the BF \& SHG test set together with some of its CoMIRs. Fig.~\ref{fig:good}, resp. Fig.~\ref{fig:bad},  shows the representations produced by the run of each method which yielded the highest, resp. lowest,  RSR over the whole test set.  
Below each CoMIR pair, the overall RSR over the entire test set for each particular run is given.

\begin{figure}[htb]
\centering
\includegraphics[width=\textwidth]{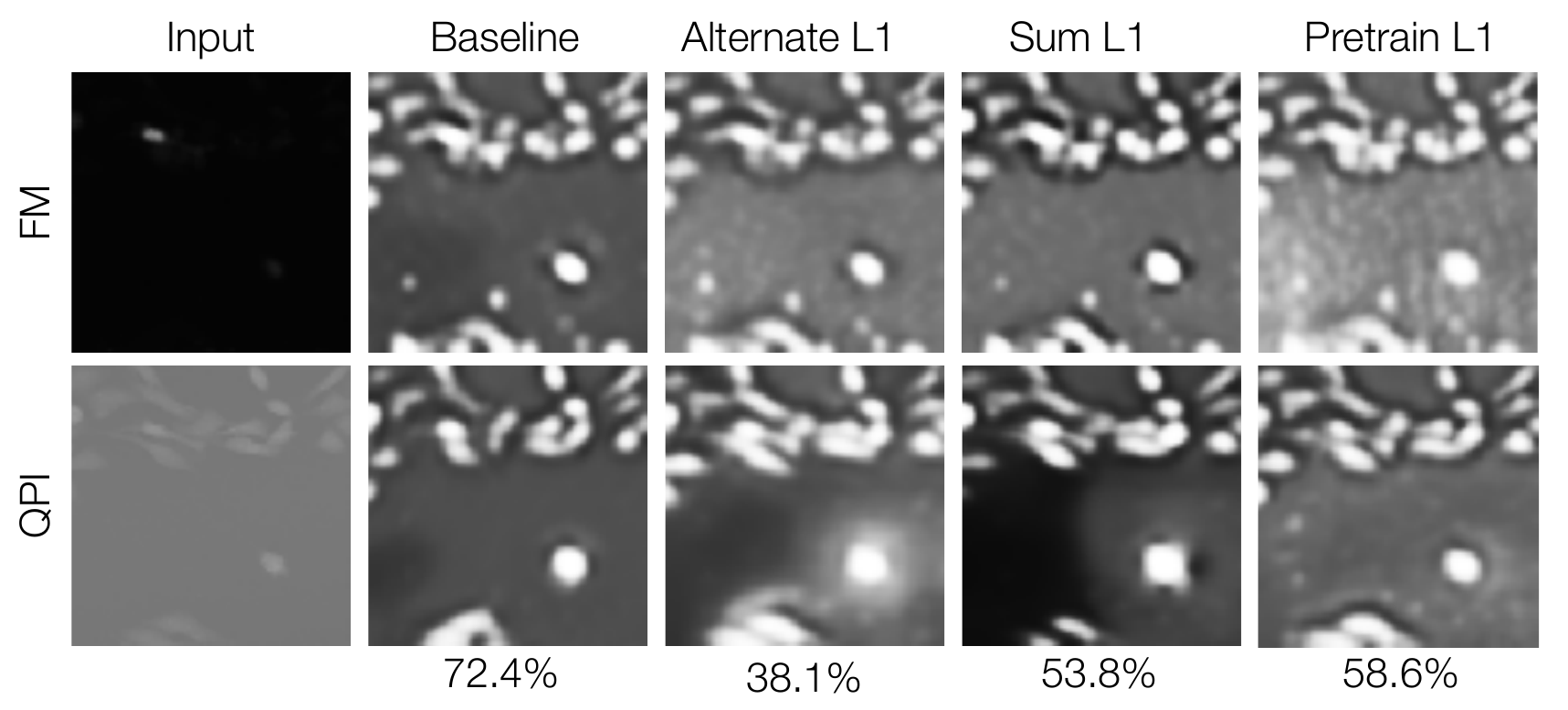}
\caption{Example of one image pair from the QPI \& FM test set and its CoMIRs produced by the studied approaches using the best performance critic in the experiments on the BF \& SHG dataset. The upper row show the FM image and its corresponding CoMIRs, the lower row show the QPI image and its CoMIRs. Below the CoMIRs, the RSR on the tlevel subset of the fold to which these CoMIRs belong is given for reference.}
\label{fig:cyto_reps}
\end{figure}

In Fig. \ref{fig:cyto_reps}, an image pair randomly selected from the QPI \& FM test set together with its CoMIRs is shown. Below each CoMIR pair, the overall RSR over the tlevel subset of the fold from which the examples images were drawn is given. 

Figure~\ref{fig:barplots} summarizes the results of evaluation of the CoMIRs generated with the different explored approaches to  supervise the BN latent representations. Fig.~\ref{fig:barplots}(a)  shows the RSR, computed as the percent of images in the test set which were successfully registered with an error less than 100\,px. Fig.~\ref{fig:barplots}(b) - (f) show the level of similarity/difference (as the measure of quality) of the generated CoMIRs, quantified by different similarity (c,d) and distance (b,e,f) measures.

\subsection{MDS Embeddings}
\begin{figure}
\centering
\includegraphics[height=\textheight]{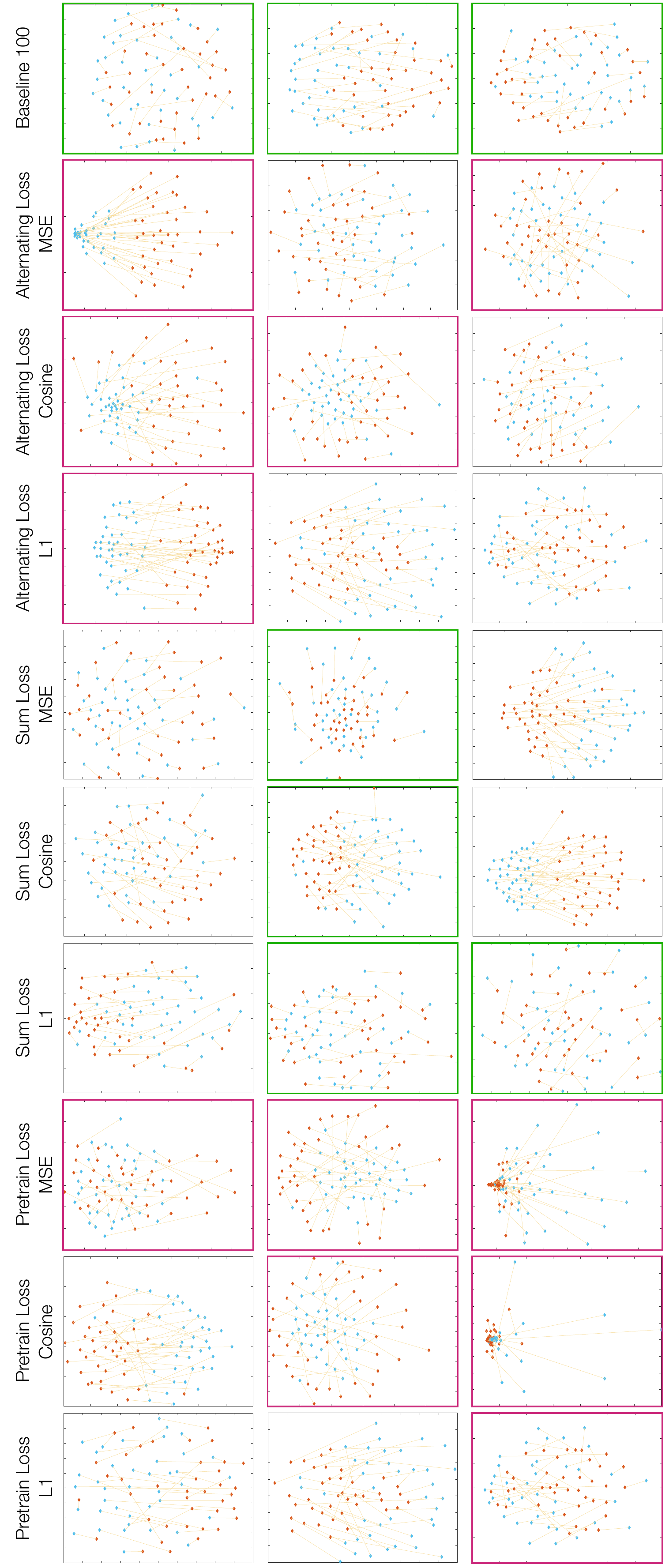}
\caption{Metric MDS solutions of CoMIRs.}
\label{fig:MDSEmbeddings2}
\end{figure}

\begin{figure}
\centering
\includegraphics[height=\textheight]{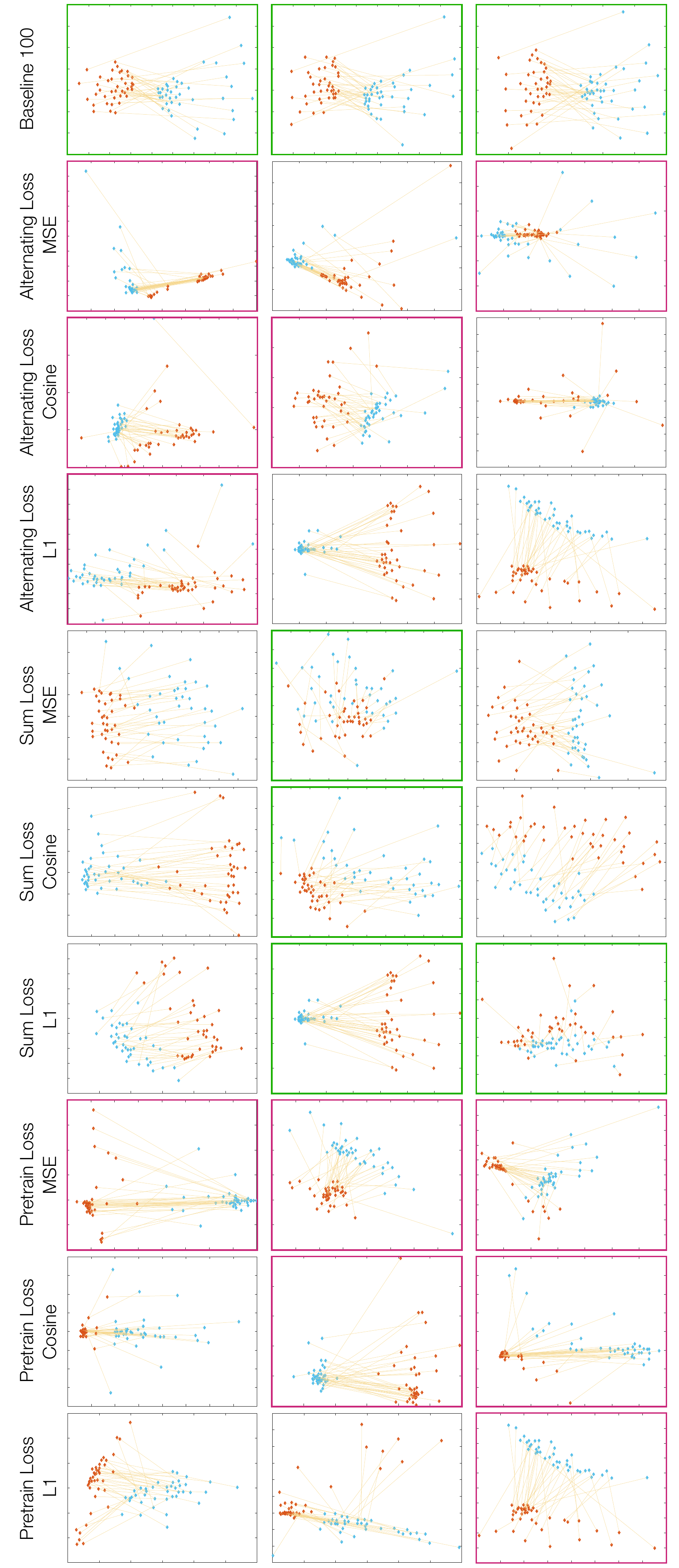}
\caption{Metric MDS solutions of BN representations.}
\label{fig:MDSEmbeddings_BN}
\end{figure}

Fig. \ref{fig:MDSEmbeddings2} shows the embeddings of CoMIR features using metric MDS with Sammon's stress criterion for all experiments. 
Fig. \ref{fig:MDSEmbeddings_BN} shows the embeddings of BN features using metric MDS with Sammon's stress criterion for all experiments. In Fig. \ref{fig:MDSEmbeddings2} and \ref{fig:MDSEmbeddings_BN} each row shows the MDS solutions for a particular approach and similarity measure configuration, the columns corresponds to the three conducted runs, corresponding to the runs shown in Table Appendix 1).
The green frames mark embeddings of runs with registration performance higher than 65\%, the magenta frames mark runs with poor registration
performance below 35\%.

\par\vfill\par

\end{document}